\documentclass[journal]{IEEEtran}
\IEEEoverridecommandlockouts
\usepackage[export]{adjustbox}
\usepackage{setspace}
\usepackage{cite}
\usepackage{amsmath,amssymb,amsfonts,mathtools,bm}
\usepackage{amsthm}
\usepackage{algorithm}
\usepackage{graphicx}
\usepackage{textcomp}
\usepackage{xcolor,float}
\usepackage{tabularx}
\usepackage{textcomp}
\usepackage{diagbox}
\usepackage{svg}
\usepackage{booktabs}
\usepackage{subfigure}
\usepackage{hyperref}

\newtheorem{problem}{Problem}
\usepackage{algpseudocode}
\usepackage{graphicx}
\usepackage{makecell}
\usepackage{amsthm}
\usepackage{caption}
\usepackage{sidecap}
\usepackage{microtype}
\usepackage{booktabs} 
\usepackage{multirow}
\usepackage{float}
\usepackage{adjustbox} 
\usepackage{amsmath}
\usepackage{amssymb}
\usepackage{colortbl}
\usepackage{makecell}
\usepackage{float}
\usepackage{url}
\usepackage{balance}
\usepackage{bbding}
\usepackage{hyperref}
\usepackage{graphicx}
\usepackage{sidecap}
\usepackage{microtype}
\usepackage{graphicx}
\usepackage{booktabs} 
\usepackage{multirow}
\usepackage{array}
\usepackage{subcaption}
\usepackage{graphicx}
\usepackage{makecell}
\usepackage{amsmath}
\usepackage{amssymb}

\newtheorem{remark}{Remark}
\usepackage{tikz}
\usepackage{changes}

\usepackage{ifthen}
\newboolean{showchanges}
\setboolean{showchanges}{false}  

\newcommand{\add}[1]{%
    \ifthenelse{\boolean{showchanges}}%
        {\textcolor{blue}{#1}}
        {#1\relax}
}

\definecolor{lime}{HTML}{A6CE39}
\DeclareRobustCommand{\orcidicon}{%
    \begin{tikzpicture}
    \draw[lime, fill=lime] (0,0) 
    circle [radius=0.16] 
    node[white] {{\fontfamily{qag}\selectfont \tiny ID}};    \draw[white, fill=white] (-0.0625,0.095) 
    circle [radius=0.007];    \end{tikzpicture}
    \hspace{-2mm}}
\foreach \x in {A, ..., Z}{%
    \expandafter\xdef\csname orcid\x\endcsname{\noexpand\href{https://orcid.org/\csname orcidauthor\x\endcsname}{\noexpand\orcidicon}}
    }

\allowdisplaybreaks
\renewcommand{\baselinestretch}{1}

\begin{document}

\title{RadioDiff-Inverse: Diffusion Enhanced Bayesian Inverse Estimation for ISAC Radio Map Construction}
\author{Xiucheng Wang\orcidA{},~\IEEEmembership{Student Member,~IEEE,}
        Zhongsheng Fang\orcidB{},~\IEEEmembership{Student Member,~IEEE,} Nan Cheng\orcidC{},~\IEEEmembership{Senior Member,~IEEE,} Ruijin Sun\orcidD{},~\IEEEmembership{Member,~IEEE}, Zan Li\orcidE{},~\IEEEmembership{Fellow,~IEEE,} and Xuemin (Sherman) Shen\orcidF{},~\IEEEmembership{Fellow,~IEEE}

\thanks{
This work was supported by the National Key Research and Development Program of China (2024YFB2907500).Xiucheng Wang, Zhongsheng Fang, Nan Cheng, Ruijin Sun, and Zan Li are with the State Key Laboratory of ISN and School of Telecommunications Engineering, Xidian University, Xi’an 710071, 
China (e-mail: xcwang\_1@stu.xidian.edu.cn; zsfang@stu.xidian.edu.cn; dr.nan.cheng@ieee.org; \{sunruijin, zanli\}@xidian.edu.cn). \textit{Corresponding author: Nan Cheng}.
Xuemin (Sherman) Shen is with the Department of Electrical and Computer Engineering, University of Waterloo, Waterloo, N2L 3G1, Canada (e-mail: sshen@uwaterloo.ca).
}        
}


\maketitle
\IEEEdisplaynontitleabstractindextext
\IEEEpeerreviewmaketitle
\begin{figure*}[t]
\captionsetup{font={small}, skip=16pt}
\scriptsize
\centering
\setlength{\tabcolsep}{2pt} 
\renewcommand{\arraystretch}{1.2} 

\begin{tabular}{l@{}*{8}{c}@{}} 
\textbf{Input } &
\includegraphics[width=0.11\linewidth, valign=m]{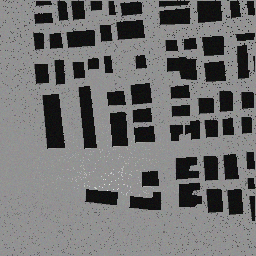} &
\includegraphics[width=0.11\linewidth, valign=m]{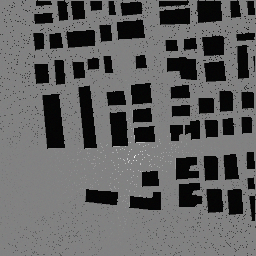} &
\includegraphics[width=0.11\linewidth, valign=m]{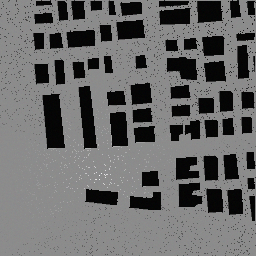} &
\includegraphics[width=0.11\linewidth, valign=m]{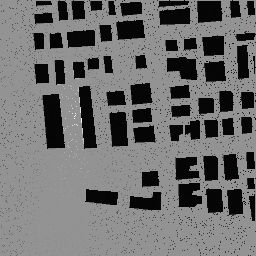} &
\includegraphics[width=0.11\linewidth, valign=m]{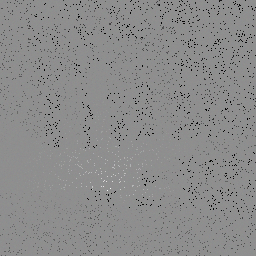} &
\includegraphics[width=0.11\linewidth, valign=m]{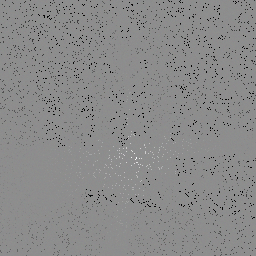} &
\includegraphics[width=0.11\linewidth, valign=m]{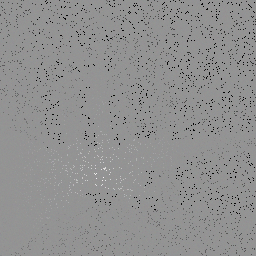} &
\includegraphics[width=0.11\linewidth, valign=m]{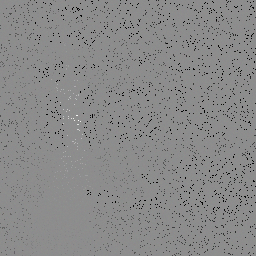} \\[3pt] 

\textbf{Output} &
\includegraphics[width=0.11\linewidth, valign=m]{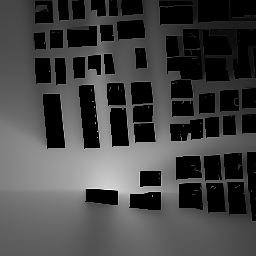} &
\includegraphics[width=0.11\linewidth, valign=m]{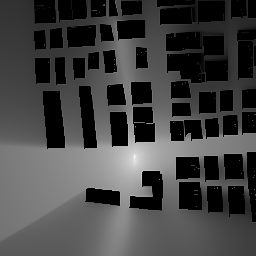} &
\includegraphics[width=0.11\linewidth, valign=m]{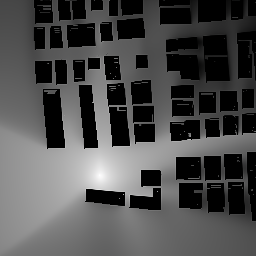} &
\includegraphics[width=0.11\linewidth, valign=m]{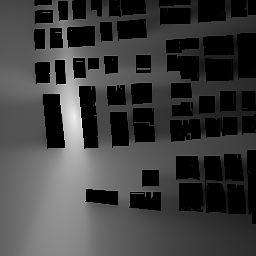} &
\includegraphics[width=0.11\linewidth, valign=m]{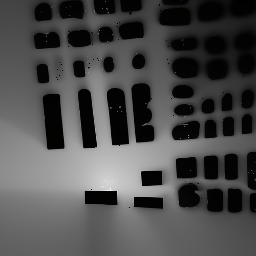} &
\includegraphics[width=0.11\linewidth, valign=m]{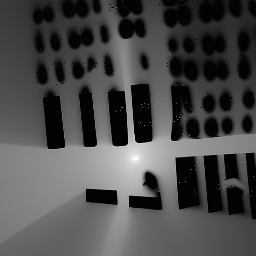} &
\includegraphics[width=0.11\linewidth, valign=m]{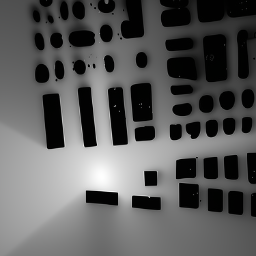} &
\includegraphics[width=0.11\linewidth, valign=m]{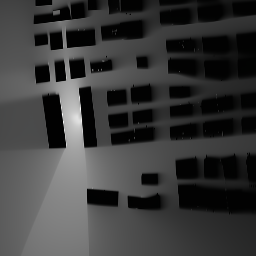} \\[3pt]

\textbf{Ground Truth} &
\includegraphics[width=0.11\linewidth, valign=m]{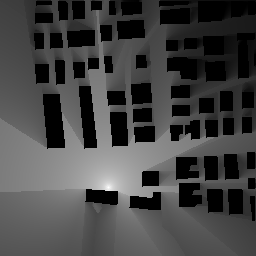} &
\includegraphics[width=0.11\linewidth, valign=m]{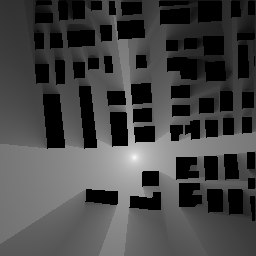} &
\includegraphics[width=0.11\linewidth, valign=m]{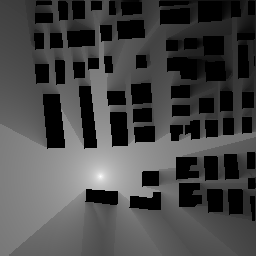} &
\includegraphics[width=0.11\linewidth, valign=m]{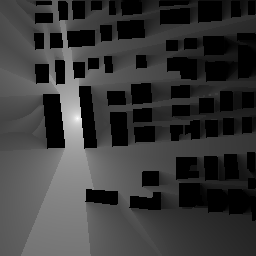} &
\includegraphics[width=0.11\linewidth, valign=m]{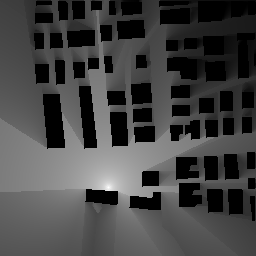} &
\includegraphics[width=0.11\linewidth, valign=m]{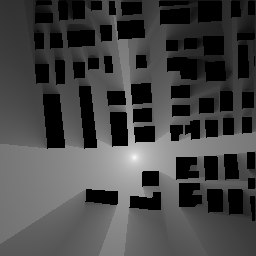} &
\includegraphics[width=0.11\linewidth, valign=m]{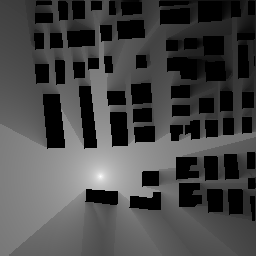} &
\includegraphics[width=0.11\linewidth, valign=m]{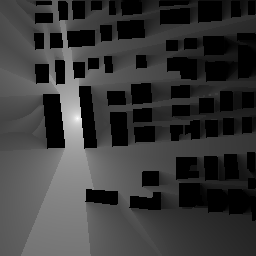} \\
\end{tabular}

\caption{Comparison of different scenarios. Row 1: Input measurements; Row 2: Output RMs; Row 3: Ground Truth .}
\label{fig:r1}
\end{figure*}
\begin{abstract}
Radio maps (RMs) are essential for environment-aware communication and sensing, providing location-specific wireless channel information. Existing RM construction methods often rely on precise environmental data and base station (BS) locations, which are not always available in dynamic or privacy-sensitive environments. While sparse measurement techniques reduce data collection, the impact of noise in sparse data on RM accuracy is not well understood. This paper addresses these challenges by formulating RM construction as a Bayesian inverse problem under coarse environmental knowledge and noisy sparse measurements. Although maximum a posteriori (MAP) filtering offers an optimal solution, it requires a precise prior distribution of the RM, which is typically unavailable. To solve this, we propose RadioDiff-Inverse, a diffusion-enhanced Bayesian inverse estimation framework that uses an unconditional generative diffusion model to learn the RM prior. This approach not only reconstructs the spatial distribution of wireless channel features but also enables environmental structure perception, such as building outlines, and location of BS just relay on pathloss, through integrated sensing and communication (ISAC). Remarkably, RadioDiff-Inverse is training-free, leveraging a pre-trained model from Imagenet without task-specific fine-tuning, which significantly reduces the training cost of using generative large model in wireless networks. Experimental results demonstrate that RadioDiff-Inverse achieves state-of-the-art performance in accuracy of RM construction and environmental reconstruction, and robustness against noisy sparse sampling.
\end{abstract}


\section{Introduction}
The evolution of wireless communication from an environment-independent paradigm to an environment-aware paradigm marks a significant shift in the design and operation of next-generation networks \cite{6g}. This transition is driven by the increasing demand for intelligent, adaptive, and context-aware communication systems, particularly in 6G networks, where network nodes must efficiently interact with their surrounding environments \cite{shen2023toward}. A key enabler of this transition is the radio map (RM), which provides a grafical representation for the spatial distribution of wireless channel characteristics, such as pathloss, time of arrival (ToA), and angle of arrival (AoA) \cite{zeng2024tutorial}. By encoding location-specific channel information, RM enables proactive network optimization and real-time adaptation to environmental changes \cite{wang2022adversarial}. One of the primary applications of RM lies in low pilot-consumption communication, where it allows infrastructure components such as intelligent reflecting surfaces (IRS) and massive multiple input multiple output (MIMO) systems to acquire high-fidelity channel state information (CSI) with no/low need for pilot signals, significantly reducing training overhead and enhancing spectral efficiency \cite{dang2020should,liu2025sensing}. Furthermore, RM plays a crucial role in trajectory planning and interference mitigation for dynamic network elements, such as autonomous aerial vehicles (AAVs) and satellites in space-air-ground integrated networks (SAGIN) \cite{cheng2019space, wang2022joint}. By precomputing the spatial distribution of wireless channel properties, RM enables mobile nodes to optimize their flight paths, resource allocation, and beamforming strategies before entering a given service region, thereby improving coverage, connectivity, and communication reliability in highly dynamic environments.

\begin{figure*}
    \centering
    \includegraphics[width=1\linewidth]{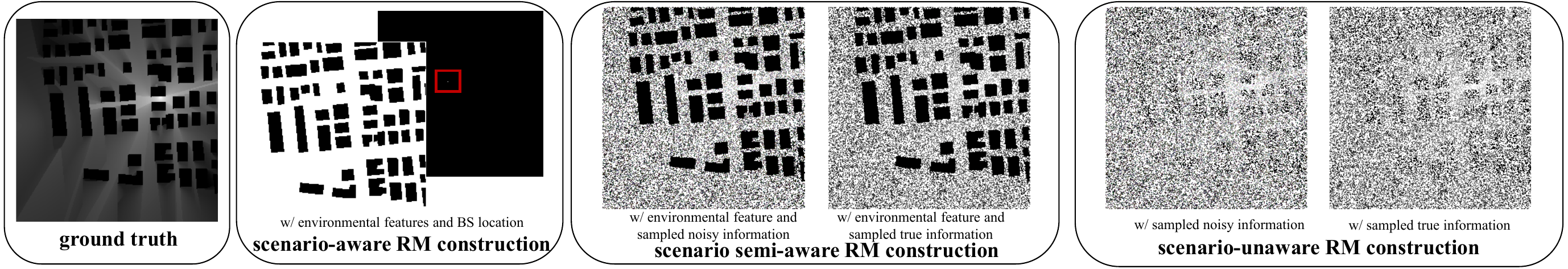}
    \caption{Illustration of RM construction under different information.}
    \label{fig-task-illustration}
    \vspace{-12pt}
\end{figure*}

Despite numerous advantages of the RM, constructing an accurate and efficient RM remains a formidable challenge, particularly in dynamic and complex environments \cite{zeng2024tutorial}. From a taxonomic perspective, RM construction can be divided into two primary categories: known-scenario RM construction \cite{levie2021radiounet,li2022radionet,wang2024radiodiff,wang2025radiodiff} and unknown or semi-aware scenario RM construction \cite{cover1967nearest, breidt2000local}, as is shown in Fig.~\ref{fig-task-illustration}. In known-scenario RM construction, all relevant environmental details, such as base station (BS) locations, the outlines and locations of environment entities are fully available, making RM construction a deterministic problem \cite{wang2024radiodiff}. In theory, given the source radiation characteristics, dielectric constants, and magnetic permeability of the environment, the propagation of electromagnetic (EM) waves at any point in space can be derived by solving Maxwell’s equations \cite{balanis2002antenna}. However, the computational complexity of directly solving Maxwell’s equations is often prohibitive, requiring high-dimensional numerical solutions \cite{bondeson2012computational}. To mitigate this, approximate methods such as electromagnetic ray tracing (ERT) have been widely adopted, providing sufficiently accurate EM wave spatial distribution estimations at the cost of limited but extensive computational time \cite{oh2004mimo}. To further enhance efficiency, neural network (NN)-based approaches have emerged as an alternative, leveraging data-driven learning to approximate RM distributions. Techniques such as UNet based RadioUNet \cite{levie2021radiounet} and Transformer-based RadioNet\cite{li2022radionet} enable fast inference by training deep models on environmental features like building layouts and BS locations. More recently, generative artificial intelligence (GAI) methods have been introduced to model RM as a stochastic process, where random Gaussian noise is incorporated during training to capture environmental uncertainty, such as generative adversarial network (GAN) based RME-GAN \cite{zhang2023rme,zhang2024fast,liao2024tfsemantic} and diffusion model (DM) based RadioDiff \cite{wang2024radiodiff} approaches.

Despite the advances in scenario-aware RM construction, its practical deployment remains severely constrained by its reliance on high-precision environmental modeling and the known of the location of BS \cite{zeng2024tutorial}. These methods assume access to detailed maps of environmental structures, which define the propagation characteristics of EM waves. However, in privacy-sensitive areas, highly dynamic urban landscapes, or adversarial environments with non-cooperative radiation sources, obtaining such high-precision environmental data is often impractical, infeasible, or restricted \cite{zhang2024radiomap}. This fundamental limitation significantly reduces the applicability of scenario-aware RM methods, particularly in real-world deployments where environmental conditions are constantly evolving. In such cases, traditional interpolation-based approaches, such as linear interpolation and Kriging, are commonly employed to estimate missing wireless channel characteristics from sparse measurements \cite{cover1967nearest,breidt2000local}. However, these methods fail to generalize in complex environments where signal propagation is highly nonlinear, due to factors such as obstructions, multipath fading, and small-scale spatial variations. 
Furthermore, prior works have largely overlooked the impact of noise in sparsely sampled data on RM construction, assuming that sparse measurements remain noise-free \cite{cover1967nearest}. In reality, noise in sparse sampling significantly degrades RM reconstruction accuracy, particularly in environments with severe multipath effects, dynamic interference, and varying signal propagation conditions \cite{levie2021radiounet}. The presence of noise in sparse measurements leads to highly unstable RM estimations, as traditional interpolation and machine learning-based methods often fail to recover structured channel characteristics under such uncertainties. Addressing scenario-unaware RM construction is particularly critical in emerging wireless applications, such as BS deployment optimization, interference management, and spectrum allocation in satellite and aerial networks \cite{zeng2024tutorial}. In these scenarios, the ability to infer the RM of non-cooperative transmitters is essential for interference avoidance, coexistence strategies, and adaptive resource allocation \cite{rose2002wireless,wang2023maploc}. However, the absence of detailed environmental models and precise BS positioning information, coupled with the challenge of noise in sparse measurements, fundamentally limits the effectiveness of traditional RM construction techniques. This limitation underscores the urgent need for alternative approaches that can both leverage sparse measurements and mitigate noise effects to accurately reconstruct RMs in complex, dynamic, and unstructured settings. To overcome these challenges, scenario-unaware and scenario semi-aware RM construction approaches offer a potential solution by bypassing the need for detailed environmental maps. Instead, these methods rely on coarse environmental knowledge and sparse wireless channel measurements to approximate the spatial distribution of EM waves. Unlike precomputed scenario-aware models, these methods provide greater adaptability in dynamic, unstructured, and privacy-sensitive environments, where environmental features may be unavailable, unreliable, or constantly evolving. However, achieving accurate RM reconstruction under sparse noisy sampling remains a fundamental challenge, necessitating the development of new techniques that can extract and infer radio propagation characteristics efficiently under uncertainty.
Moreover, an often-overlooked yet fundamental characteristic of RMs is their bidirectional relationship with the surrounding environment. While traditionally employed to represent the spatial distribution of wireless channel characteristics, RMs inherently encode structural information about their environment. Variations in pathloss, ToA, and AoA across space provide implicit cues about the geometry, material properties, and layout of surrounding objects. This intrinsic property extends RM’s utility beyond mere communication optimization, making it a key enabler of integrated sensing and communication (ISAC), where wireless communication and environmental perception are seamlessly integrated \cite{liu2022survey}. However, leveraging RMs for environment inference and environment unware RM construction poses a significant challenge, as it necessitates a robust framework capable of extracting meaningful prior knowledge from coarse environmental information and sparse, noisy wireless measurements. To address these challenges, we formulate RM construction as a Bayesian inverse problem, where the objective is to infer the underlying RM distribution from limited and noisy observations. Bayesian inference provides a principled framework for handling uncertainty and incorporating prior knowledge to improve estimation accuracy. Although maximum a posteriori (MAP) filtering offers a theoretically optimal solution to the Bayesian inverse problem, it requires an accurate prior distribution of the RM, which is often unavailable or difficult to model explicitly \cite{box2011bayesian}. To circumvent this limitation, DMs—which have demonstrated remarkable generative capabilities—offer a promising alternative by learning the prior distribution of RMs across diverse environments. By integrating diffusion models into the Bayesian inference framework, we can construct RMs even when precise BS locations and detailed environmental maps are missing, while simultaneously mitigating the adverse effects of noise in sparse measurements. To this end, we propose RadioDiff-Inverse, a diffusion-enhanced Bayesian inverse framework designed for robust RM reconstruction under coarse environmental knowledge and sparse, noisy measurements. Unlike traditional approaches that rely on deterministic interpolation or predefined environmental models, RadioDiff-Inverse leverages diffusion-based generative modeling to learn the underlying statistical structure of RMs, enabling adaptive RM reconstruction in highly dynamic and uncertain environments. By integrating Bayesian inverse estimation with diffusion-based generative modeling, RadioDiff-Inverse represents a paradigm shift in RM construction, moving beyond static scenario-aware methods to a data-driven, uncertainty-aware framework. The resulting RM not only reconstructs the spatial distribution of wireless channel features but also enables accurate environmental inference—such as detecting the outlines and locations of buildings and obstacles—through the ISAC framework. As a result, RadioDiff-Inverse provides a powerful, scalable solution for next-generation wireless communication systems, particularly in dynamic, unstructured, and privacy-sensitive environments where traditional RM construction methods fail. The main contributions of this paper are summarized as follows.
\begin{enumerate}
    \item To address the challenges of scenario unware/semi-aware RM construction, we formulate RM estimation as a Bayesian inverse problem and propose RadioDiff-Inverse, a novel framework that employs maximum a posteriori (MAP) filtering to reconstruct RM under sparse and noisy sampling conditions. By treating RM construction as an inverse problem, our approach enables the inference of missing environmental and BS information, allowing for RM reconstruction in dynamic, unstructured, and privacy-sensitive environments. Unlike traditional approaches that rely on deterministic interpolation or predefined environmental models, RadioDiff-Inverse introduces a probabilistic inference framework, significantly enhancing the robustness of RM reconstruction under uncertainty.
    \item To effectively solve this Bayesian inverse problem, we develop an unconditional generative DM, pretraind on Imagenet without any task-specific fine-tune or post-training to extract the prior distribution of RM across diverse environments. By leveraging the intrinsic properties of diffusion-based generative modeling, our approach captures the underlying statistical patterns of RM, even when precise environmental details are unavailable.
    \item Through extensive experimental validation, we demonstrate that RadioDiff-Inverse achieves state-of-the-art (SOTA) performance in RM reconstruction accuracy, environmental perception, BS localization just based on pathloss, and robustness to sparse noisy sampling. Our results show that the proposed method can accurately reconstruct spatial distributions of wireless channel characteristics while simultaneously inferring environmental structures, a capability that is crucial for ISAC applications. Furthermore, RadioDiff-Inverse significantly outperforms traditional interpolation methods and deterministic generative models in scenarios where high-precision environmental information is unavailable, reinforcing its applicability in next-generation wireless networks, autonomous systems, and satellite communication infrastructures.
\end{enumerate}

\section{Preliminary}
\subsection{Bayesian inverse problems}
Bayesian inverse problems arise in numerous scientific and engineering applications where one seeks to recover an unknown signal or parameter from indirect and often noisy observations \cite{box2011bayesian}. These problems are particularly prevalent in fields such as medical imaging, geophysics, computer vision, and wireless signal processing, where direct measurements of the desired quantities are either infeasible or incomplete. Mathematically, an inverse problem is typically formulated as a system of equations that relate the unknown variable $\bm{x}$ to the observed data $\bm{y}$ through a known forward operator $A$, often corrupted by noise as follows.
\begin{align}
\bm{y} = A \bm{x} + \bm{n}, \quad \bm{n} \sim \mathcal{N}(0, \sigma^2 I).
\end{align}
where $\bm{x} \in \mathbb{R}^{D}$ represents the unknown quantity to be estimated, $\bm{y} \in \mathbb{R}^{d}$ denotes the observed measurements, and $A \in \mathbb{R}^{d \times D}$ is a known transformation matrix encoding the mapping between the two. The term $\bm{n}$ represents measurement noise, which is commonly assumed to follow a Gaussian distribution with variance $\sigma^2$. A fundamental challenge in inverse problems arises when the number of observations is smaller than the number of unknowns (i.e., $d < D$), making the system underdetermined and the solution non-unique. Such problems are often classified as ill-posed, necessitating additional constraints or regularization techniques to obtain a meaningful estimate of $\bm{x}$.

A principled approach to addressing the ill-posed nature of inverse problems is to adopt a Bayesian inference framework, which incorporates prior knowledge about the unknown variable $\bm{x}$. Instead of seeking a single deterministic solution, the Bayesian approach formulates the inverse problem as an inference task, where the goal is to estimate the posterior distribution of $\bm{x}$ given the observations $\bm{y}$. Using Bayes’ theorem, the posterior distribution is expressed as follows.
\begin{align}
p(\bm{x} | \bm{y}) \propto p(\bm{x}) p(\bm{y} | \bm{x}).
\end{align}
where $p(\bm{x})$ represents the prior distribution, which encodes prior knowledge about the likely values of $\bm{x}$ based on domain-specific assumptions. The term $p(\bm{y} | \bm{x})$ is the likelihood function, which models the probability of observing $\bm{y}$ given a specific realization of $\bm{x}$. Under the assumption of Gaussian noise in the measurement process, the likelihood follows a normal distribution as follows.
\begin{align}
p(\bm{y} | \bm{x}) = \mathcal{N}(\bm{y} | A \bm{x}, \sigma^2 I).
\end{align}
By combining the prior with the likelihood, Bayesian inference provides a posterior distribution that quantifies the uncertainty in the recovered solution. This formulation naturally leads to several inference strategies for estimating $\bm{x}$. One common approach is Maximum a Posteriori (MAP) estimation, which seeks the most probable value of $\bm{x}$ by solving:
\begin{align}
\hat{\bm{x}}_{\text{MAP}} = \arg \max_{\bm{x}} p(\bm{x} | \bm{y}) = \arg \max_{\bm{x}} p(\bm{x}) p(\bm{y} | \bm{x}).
\end{align}
In cases where the prior is Gaussian, i.e., $p(\bm{x}) = \mathcal{N}(\bm{x} | 0, \Sigma)$, the MAP estimate reduces to a regularized least-squares problem as follows.
\begin{align}
\hat{\bm{x}}_{\text{MAP}} = \left( A^T A + \sigma^2 \Sigma^{-1} \right)^{-1} A^T \bm{y},
\end{align}
which corresponds to Tikhonov regularization in classical inverse problem literature.

\subsection{Score-Based Denoising Diffusion Model}\label{sec-2-b}
\begin{figure*}[ht]
    \centering
    \includegraphics[width=1\linewidth]{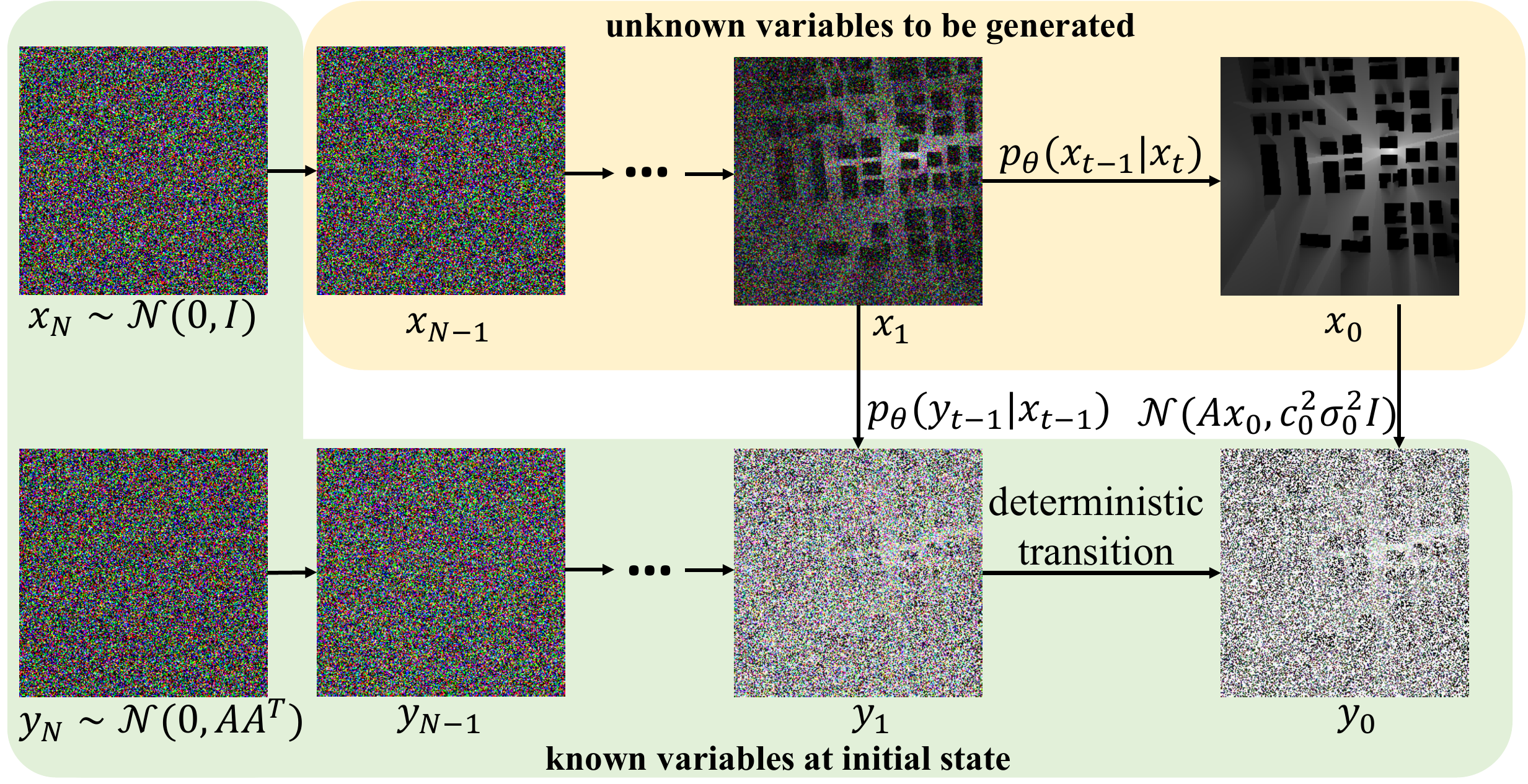}
    \caption{The illustration of inferencing sequence.}
    \label{fig-inference-sequence}
    \vspace{-12pt}
\end{figure*}
DM have emerged as a powerful class of generative models capable of producing high-quality samples from complex distributions \cite{ho2020denoising}. Among them, score-based diffusion models provide a probabilistic framework in which data is gradually perturbed using a stochastic process, and the generative model learns to reverse this process through score function estimation \cite{song2020score}. Unlike standard DMs, which rely on discrete Markov chains for noise injection and denoising, score-based diffusion models utilize stochastic differential equations (SDEs) to continuously perturb and reconstruct data distributions. This formulation naturally aligns with Bayesian inference, enabling efficient posterior sampling for inverse problems, such as radio map reconstruction. Formally, the forward diffusion process is defined as a stochastic perturbation process, where an initial data distribution $p_0(\bm{x})$ is gradually transformed into a known prior, typically as Gaussian, via a controlled SDE as follows.
\begin{align}
d\bm{x} = f(\bm{x}, t) dt + g(t) d\bm{w},
\end{align}
where $f(\bm{x}, t)$ is the drift term that governs deterministic transformation, $g(t)$ is a time-dependent diffusion coefficient, and $d\bm{w}$ represents a standard Wiener process. As $t$ progresses from $0$ to $T$, the distribution $p_t(\bm{x})$ transitions from the data distribution to an isotropic Gaussian distribution, ensuring tractable sampling. The generative process in score-based diffusion models relies on reversing the forward stochastic process, which involves solving a corresponding reverse-time SDE as follows.
\begin{align}
d\bm{x} = \left[ f(\bm{x}, t) - g^2(t) \nabla_{\bm{x}} \log p_t(\bm{x}) \right] dt + g(t) d\bm{\bar{w}},
\end{align}
where $\nabla_{\bm{x}} \log p_t(\bm{x})$ is the score function, representing the gradient of the log probability density at time $t$, and $d\bm{\bar{w}}$ is a standard Wiener process in reverse time. In practice, $p_t(\bm{x})$ is unknown, so a neural network $s_\theta(\bm{x}, t)$ is trained to approximate the score function as follows.
\begin{align}
s_\theta(\bm{x}, t) \approx \nabla_{\bm{x}} \log p_t(\bm{x}).
\end{align}
By leveraging score-matching objectives, the model learns to recover data from noise, effectively inverting the stochastic diffusion process.

Although the reverse-time SDE formulation provides a stochastic sampling procedure, an equivalent probability flow ordinary differential equation (ODE) can be derived as follows.
\begin{align}
d\bm{x} = \left[ f(\bm{x}, t) - \frac{1}{2} g^2(t) \nabla_{\bm{x}} \log p_t(\bm{x}) \right] dt.
\end{align}
This ODE formulation eliminates stochasticity in sampling, enabling deterministic inference paths akin to denoising diffusion probabilistic models (DDPMs) \cite{ho2020denoising}. Specifically, DDPMs discretize the forward diffusion process into a finite sequence of steps as follows.
\begin{align}
q(\bm{x}_t | \bm{x}_{t-1}) = \mathcal{N}(\bm{x}_t; \alpha_t \bm{x}_{t-1}, \beta_t \bm{I}),
\end{align}
where $\alpha_t$ and $\beta_t$ are step-wise coefficients controlling noise injection. The generative process then reconstructs data using a learned noise estimator $\epsilon_\theta(\bm{x}_t, t)$, which corresponds to the score function as follows.
\begin{align}
s_\theta(\bm{x}, t) = -\frac{\epsilon_\theta(\bm{x}, t)}{\sqrt{1 - \bar{\alpha}_t}}.
\end{align}
Thus, DDPM can be interpreted as a discrete implementation of score-based diffusion using a variance-preserving diffusion process.

The decoupled diffusion model (DDM) \cite{huang2024decoupled}, which is employed in the SOTA scenario-aware RM construction method RadioDiff, introduces a structured diffusion framework by decoupling data attenuation and noise injection into two distinct stages. Unlike conventional diffusion models that directly introduce Gaussian noise to the original data, DDM first attenuates the initial state $\bm{n}_0$ to a zero vector before adding stochastic noise. This forward process is defined as a continuous Markov process, where the transition from $\bm{n}_0$ to $\bm{n}_t$ follows a Gaussian distribution as follows.
\begin{align}
q\left(\bm{x}_t \mid \bm{x}_0\right) = \mathcal{N}\left(\gamma_t \bm{x}_0, \delta_t^2 \bm{I}\right),
\end{align}
where $\gamma_t$ and $\delta_t$ as time-dependent coefficients controlling the attenuation and noise variance, respectively. The differential form of this process follows.
\begin{align}
&d \bm{x}_t = f_t \bm{x}_t dt + g_t d\bm{\epsilon}_t,\\
&f_t = \frac{d \log \gamma_t}{dt},\\
&g_t^2 = \frac{d \delta_t^2}{dt} - 2 f_t \delta_t^2
\end{align}
where $f_t$ governs the rate of decay and $g_t^2$ determines the noise accumulation. This structured perturbation stabilizes the diffusion process, mitigating early-stage variance and ensuring more controlled generative dynamics.
The reverse process reconstructs $\bm{x}_0$ from $\bm{x}_t$ by solving the corresponding stochastic differential equation as follows.
\begin{align}
d \bm{x}_t = \left[f_t \bm{x}_t - g_t^2 \nabla_{\bm{x}} \log q\left(\bm{x}_t\right)\right] dt + g_t d\overline{\bm{\epsilon}}_t.
\end{align}
By leveraging the decoupling property, DDM enhances stability in both training and inference, ensuring improved sample quality and reduced computational overhead. The transformation from $\bm{x}_0$ to $\bm{0}$ is deterministic, simplifying the forward process as follows.
\begin{align}
q(\bm{x}_t|\bm{x}_0) = \mathcal{N}\left(\bm{x}_{0}+\int_{0}^{t}\bm{f}_t\mathrm{d}t,t\bm{I}\right),
\end{align}
which leads to an efficient reverse sampling scheme as follows.
\begin{align}
    q\left(\bm{x}_{t-\Delta t} \mid \bm{x}_t, \bm{x}_0\right)  &=\mathcal{N}\left(\bm{x}_{t} +\int_t^{t-\Delta t} \bm{f}_t \mathrm{~d} t\right. \notag\\
& \left.\qquad\qquad-\frac{\Delta t}{\sqrt{t}} \bm{\epsilon}_{t}, \frac{\Delta t(t-\Delta t)}{t} \bm{I}\right).\label{ddm-reverse}
\end{align}

\section{Problem Formulation}
In this work, we consider the task where the RM needs to be constructed in a given area, represented as a grid of size \(N \times N\). Each grid cell is assumed to have a constant pathloss value, allowing the RM to be expressed as a pathloss matrix $\bm{P} \in \mathbb{R}^{N \times N}$. Unlike traditional scenario-aware RM construction, we operate under scenario semi-aware conditions, meaning that precise BS location information is unavailable, while only coarse environmental knowledge and sparse, noisy measurements are provided. Within this region, there exists an unknown BS, whose location, denoted as $\bm{R}$, is not explicitly given. However, the environment contains some obstacles, which affect the propagation of EM waves. The environmetal obstacles, such as buildings and walls, have fixed positions, varying sizes and shapes, and are composed of materials that reflect, absorb, and diffract EM waves. Similar to prior works \cite{zhang2023rme, levie2021radiounet, li2022radionet}, the pathloss within their interiors is assumed to be zero. The presence of static obstacles is represented by a matrix $\bm{h}_s \in \mathbb{R}^{N \times N}$, where $h_{i,j}^{s} = 0, \forall h_{i,j}^{s} \in \bm{h}_s$ indicates the absence of a static obstacle at grid cell $(i,j)$. Unlike fully scenario-aware RM construction, where large-scale environmental features are available, we assume only sparse and noisy pathloss measurements can be obtained. Let $\bm{y}$ represent the sparse sampled measurements of the true pathloss matrix $\bm{p}$, with noise $\bm{n}$ modeled as an additive term as follows.
\begin{align}
\bm{y} = \mathcal{S}(\bm{p}) + \bm{n},
\end{align}
where $\mathcal{S}(\cdot)$ is a sampling operator that selects a small subset of the total RM points, and $\bm{n} \sim \mathcal{N}(0, \sigma^2)$ represents measurement noise. The objective is to train a NN $\bm{\mu}_{\bm{\theta}}(\cdot)$ with parameters $\bm{\theta}$ to predict the pathloss matrix $\bm{\hat{p}}$ using coarse environmental information and sparse, noisy pathloss measurements while not relying on BS location information. The reconstruction accuracy is measured using a criterion function $\mathcal{L}(\hat{\bm{p}}, \bm{p})$, which quantifies the difference between the predicted $\bm{\hat{p}}$ and the ground truth $\bm{p}$. The scenario semi-aware RM construction problem can thus be formulated as follows
\begin{problem}\label{p2}
    \begin{align}
    &\min_{\bm{\theta}}&&\mathcal{L}(\hat{\bm{p}}, \bm{p}),\label{obj}\\
    &s.t. &&\hat{\bm{P}}=\bm{\mu}_{\bm{\theta}}(\bm{y})\tag{\ref{obj}a},
\end{align}
\end{problem}
This formulation captures the key challenges of scenario semi-aware RM construction, where no explicit BS location information is available, only coarse environmental features are provided, and sparse noisy measurements must be leveraged to accurately reconstruct the RM.

In the context of RM construction under the condition of known partial environmental information and noisy sparse sampling, we aim to reconstruct an RM based on limited, noisy observations. The RM is a grid representation of wireless channel characteristics, which can be flattened into a 1-dimensional vector $\bm{x} \in \mathbb{R}^{N^2}$, where $N$ represents the grid size. Each element of $\bm{x}$ corresponds to a pathloss value or another characteristic (e.g., time of arrival, angle of arrival) at a specific location in the grid. However, due to practical constraints, only partial environmental information is available for constructing the RM. Additionally, the data we receive is sparse, meaning that only a subset of the grid cells are measured, and these measurements come with noise. This situation leads to a problem that can be formulated as a Bayesian inverse problem of the form as follows.
\begin{align}
\bm{y} = A\bm{x} + \bm{n},
\end{align}
where $\bm{y} \in \mathbb{R}^d$ is the observed data vector, which includes both the partial environmental information (e.g., coarse grid features) and noisy sparse sampling information. The data $\bm{y}$ is a sparse and noisy version of the true RM. $\bm{n} \in \mathbb{R}^d$ is the noise vector in the sparse sampling measurements, assumed to follow a Gaussian distribution $\bm{n} \sim \mathcal{N}(0, \sigma^2 I)$. $A \in \mathbb{R}^{d \times N^2}$ is the mask matrix, where only the elements on the diagonal are 1, and all other elements are 0. The mask matrix represents the sampling pattern. If the diagonal element $A_{i,i} = 1$, it indicates that the corresponding element $x_i$ of the flattened RM has been observed and is not masked; otherwise, if $A_{i,i} = 0$, the corresponding element $x_i$ has been masked and is unavailable in the measurement. In this setup, we have a sparse and noisy observation $\bm{y}$, and our goal is to infer the full RM vector $\bm{x}$ from these incomplete measurements. The Bayesian inverse problem is the task of inferring $\bm{x}$ given the observed data $\bm{y}$, the known mask matrix $A$, and the noise model. We express this as a posterior distribution as follows.
\begin{align}
p(\bm{x} | \bm{y}) \propto p(\bm{x}) p(\bm{y} | \bm{x}),
\end{align}
where $p(\bm{x})$ is the prior distribution that encodes the statistical properties or assumptions about the RM structure (e.g., smoothness or spatial correlation in wireless channels). $p(\bm{y} | \bm{x})$ is the likelihood function, which models the likelihood of observing $\bm{y}$ given a particular realization of $\bm{x}$, governed by the noise model $\bm{y} = A\bm{x} + \bm{n}$. Since the noise is assumed to be Gaussian, the likelihood is expressed as follows.
\begin{align}
p(\bm{y} | \bm{x}) = \mathcal{N}(\bm{y} | A\bm{x}, \sigma^2 I).
\end{align}
This problem formulation illustrates how RM construction under coarse environmental knowledge, sparse sampling, and measurement noise can be framed as a Bayesian inverse problem. The goal is to estimate the full RM vector $\bm{x}$ while accounting for sparse, noisy data and incomplete information, thereby enabling a more robust and accurate RM reconstruction in challenging environments where traditional methods may fail due to data sparsity or noise.

\section{Diffusion Enhanced Bayesian Inverse Estimation}
In this section, we give a detailed introduction for the framework of the proposed RadioDiff-Inverse method.
\subsection{Bayesian Filtering}
\begin{figure*}
    \centering
    \includegraphics[width=0.75\linewidth]{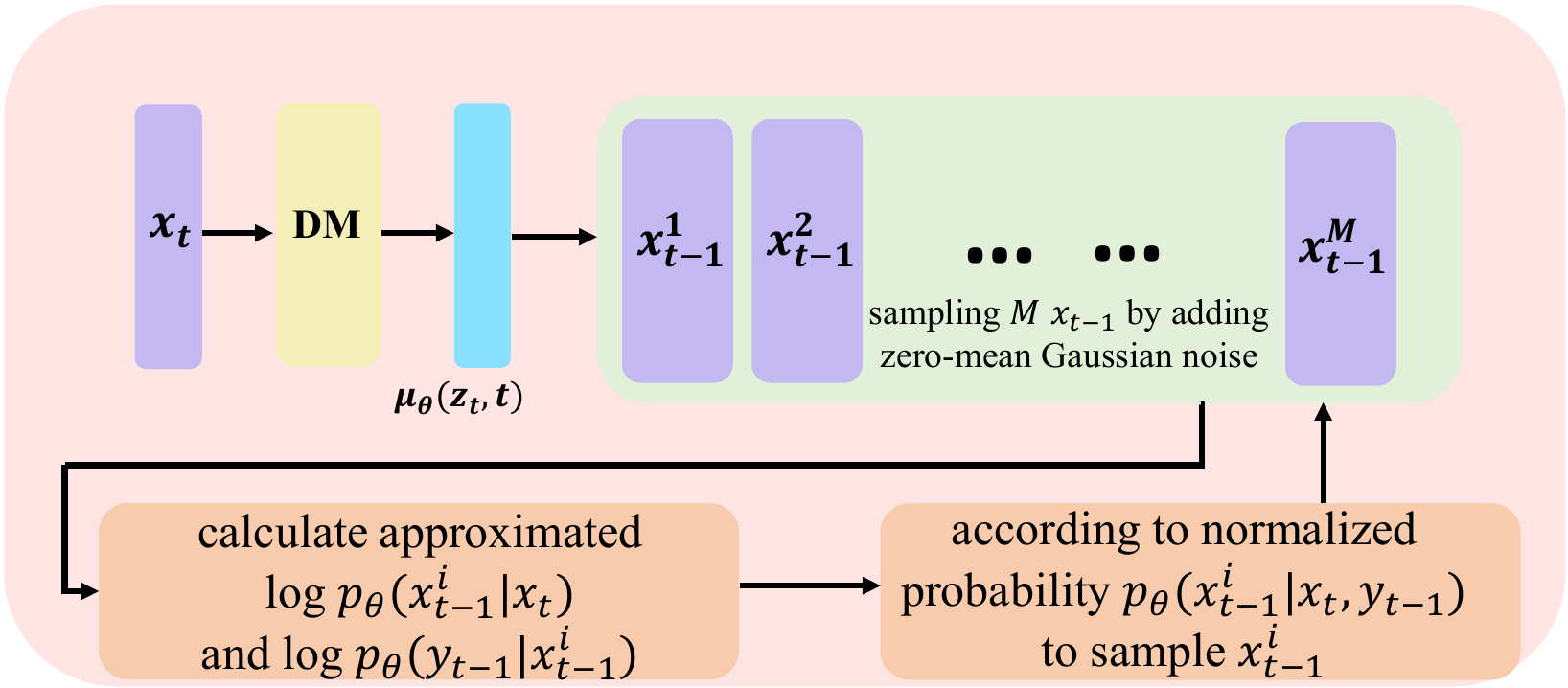}
    \caption{The illustration of denoising procedure of RadioDiff-Inverse.}
    \label{fig_sampling}
\end{figure*}
Bayesian filtering provides a framework for recursively updating the belief about a latent variable $\bm{x}_t$ based on new observations $\bm{y}_t$. The core of Bayesian filtering consists of two main steps: prediction and update. These steps, when applied iteratively over time, allow for the estimation of the posterior distribution $p(\bm{x}_t | \bm{y}_{1:k})$ at each time step $k$. The prediction step leverages the prior knowledge of how the state evolves over time, while the update step incorporates new observations to adjust the prediction and refine the estimate of the state. The prediction step is mathematically formulated as follows.
\begin{align}
p(\bm{x}_t | \bm{y}_{1:k-1}) = \int p(\bm{x}_t | \bm{x}_{t-1}) p(\bm{x}_{t-1} | \bm{y}_{1:k-1}) d\bm{x}_{t-1},
\end{align}
where $p(\bm{x}_t | \bm{x}_{t-1})$ represents the transition model that governs how the state evolves from $\bm{x}_{t-1}$ to $\bm{x}_t$, and $p(\bm{x}_{t-1} | \bm{y}_{1:k-1})$ is the prior distribution of the state at time $k-1$. The update step incorporates the new observation $\bm{y}_t$ to adjust the predicted distribution using Bayes' rule:

\begin{align}
p(\bm{x}_t | \bm{y}_{1:k}) = \frac{p(\bm{y}_t | \bm{x}_t) p(\bm{x}_t | \bm{y}_{1:k-1})}{\int p(\bm{y}_t|\bm{x}_t)p(\bm{y}_t | \bm{y}_{1:k-1})d \bm{x}_{t}},
\end{align}
where $p(\bm{y}_t | \bm{x}_t)$ is the likelihood, representing the probability of observing $\bm{y}_t$ given the state $\bm{x}_t$, and $p(\bm{y}_t | \bm{y}_{1:k-1})$ is the normalization term to ensure that the posterior is a valid probability distribution.

Repeatedly executing these two steps allows for Bayesian filtering to solve the Bayesian inverse problem. For systems where both the dynamics and measurements follow a linear Gaussian distribution, the Kalman filter can effectively solve this problem. However, for more complex, nonlinear problems, a more sophisticated method is required to estimate the posterior distribution. In this work, we propose utilizing DMs to extract prior distribution characteristics, which enables the use of Bayesian posterior sampling in a nonlinear setting.

\subsection{Diffusion Models for Bayesian Filtering}
DMs provide a natural method for solving Bayesian filtering problems. These models involve a forward diffusion process that gradually adds noise to the data and a backward sampling process that generates posterior samples. In the context of Bayesian filtering, the forward diffusion process corresponds to the prediction step, while the backward sampling process corresponds to the update step. The forward diffusion process can be expressed as follows.
\begin{align}
\bm{x}_t = a_t \bm{x}_{t-1} + b_t \bm{n}_t,
\end{align}
where $\bm{n}_t$ is independent standard normal noise, and $a_t$ and $b_t$ are scaling factors for the state and the noise. According to \cite{song2021solving, dou2024diffusion}, the noise sharing method is used as the diffusion process in the sequence of $\bm{y}$ as follows.
\begin{align}
    \bm{y}_{t} = a_{k}\bm{y}_{t-1}+b_{t}\bm{An}_{t}.
\end{align}
Therefore since the $\bm{y}_{0}=\bm{y}\sim\mathcal{N}(\bm{Ax}_{0},\sigma^2\bm{I})$, the observation model is similarly affected by noise, leading to the noisy observations as follows.
\begin{align}
\bm{y}_t \sim \mathcal{N}(\bm{A} \bm{x}_t, c_t^2 \sigma_{t}^2 \bm{I}),\label{y-from-x}
\end{align}
where $c_t = a_1 a_2 \cdots a_t$, and $\sigma^2$ represents the noise variance. This formulation establishes the Bayesian posterior sampling problem as a reverse-time Bayesian filtering problem, which is well-studied with existing algorithms and solutions. By following the prediction and update steps in Bayesian filtering, we compute the marginal posterior distribution $p_{\theta}(\bm{x}_t | \bm{y}_{t: N})$, which enables us to recover the initial latent state $\bm{x}_0$ from the noisy measurements.

\subsubsection{Generation Sequence}\label{section4-2-1}
To generate the sequence of states $\bm{x}_t$, we need to obtain the sequence $\bm{y}_{t:N}$. Although we have full access to the initial noisy measurement $\bm{y}_0$, which represents the coarse environmental features and noisy sparse measurements, this alone is not enough for posterior sampling according to Bayesian filtering principles. Each observation $\bm{y}_t$ should be sampled from a Gaussian distribution with the mean defined by $\bm{A} \bm{x}_t$. 

Given this, we use the backward sampling process to sample $\bm{y}_t$ from the sequence $\bm{y}_{t+1}$. This backward process allows us to sample from the posterior distribution $p_{\theta}(\bm{y}_t | \bm{x}_t)$. As stated in \cite{dou2024diffusion}, the equation for sampling $\bm{y}_t$ is as follows.
\begin{align}
\bm{y}_t = u_t \bm{y}_0 + v_t \bm{y}_t + w_t \cdot \bm{A} \bm{n}_t,
\end{align}
where $u_t$, $v_t$, and $w_t$ are determined by the diffusion parameters. This process is computed recursively, where the initial point $\bm{y}_N$ is sampled from $\mathcal{N}(0, \bm{A} \bm{A}^T)$, and subsequent values are generated based on the prior $\bm{y}_0$.

\subsubsection{Backward Sequence}\label{sec-4-2-2}
The backward sampling sequence is crucial for recursively sampling $\bm{x}_t$ from the noisy observations. Given that $\bm{x}_N$ is approximately a standard Gaussian, we can express the posterior distribution of $\bm{x}_N$ given the observations $\bm{y}_N$ as follows.
\begin{align}
\bm{x}_N \sim p_{\theta}(\bm{x}_N | \bm{y}_N) \propto p_{\theta}(\bm{x}_N) \cdot p_{\theta}(\bm{y}_N | \bm{x}_N).
\end{align}
The posterior distribution $p_{\bm{\theta}}(\bm{x}_N | \bm{y}_N)$ can be derived from the likelihood $p_{\bm{\theta}}(\bm{y}_N | \bm{x}_N)$, which is expressed as follows.
\begin{align}
p_{\bm{\theta}}(\bm{y}_N | \bm{x}_N) = q(\bm{y}_N | \bm{x}_N) \mathcal{N}(\bm{A} \bm{x}_N, c_N^2 \sigma^2 \bm{I}),
\end{align}
where $c_N = a_1 a_2 \cdots a_N$. The posterior distribution $p_{\bm{\theta}}(\bm{x}_N | \bm{y}_N)$ is Gaussian and can therefore be expressed in closed form. Next, we recursively sample $\bm{x}_{t-1}$ conditioned on $\bm{x}_t$ and $\bm{y}_{t-1}$. The likelihood $p_{\bm{\theta}}(\bm{y}_{t-1} | \bm{x}_{t-1})$ is given as follows.
\begin{align}
p_{\bm{\theta}}(\bm{y}_{t-1} | \bm{x}_{t-1}) = q(\bm{y}_{t-1} | \bm{x}_{t-1}) = \mathcal{N}(\bm{A} \bm{x}_{t-1}, c_{t-1}^2 \sigma^2 \bm{I}),
\end{align}
where $c_{t-1} = a_1 a_2 \cdots a_{t-1}$. Additionally, the posterior $p_{\bm{\theta}}(\bm{x}_{t-1} | \bm{x}_t)$ is determined by the score function as follows.
\begin{align}
p_{\bm{\theta}}(\bm{x}_{t-1} | \bm{x}_t) = \mathcal{N} \left( u_t \hat{\bm{x}}_0(\bm{x}_t) + v_t \bm{s}_{\bm{\theta}}(\bm{x}_t, t_t), w_t^2 \bm{I} \right),
\end{align}
where $\hat{\bm{x}}_0(\bm{x}_t) := \frac{\bm{x}_t + d_t^2 \bm{s}_{\bm{\theta}}(\bm{x}_t, t_t)}{c_t}$ is the conditional expectation of $\bm{x}_0$ given $\bm{x}_t$, computed using Tweedie’s formula. The parameters $c_t$, $d_t$, $u_t$, $v_t$, and $w_t$ are determined through the process of unconditional diffusion sampling. Specific values for these parameters in the DDPM and DDM frameworks are provided in Section \ref{sec-2-b}.
Now, we can compute the posterior distribution as follows.
\begin{align}
p_{\bm{\theta}}(\bm{x}_{t-1} | \bm{x}_t, \bm{y}_{t-1}) &=\frac{p_{\bm{\theta}}\left(\bm{x}_{t-1},\bm{y}_{t-1}|\bm{x}_{t}\right)}{p_{\bm{\theta}}\left(\bm{y}_{t-1}|\bm{x}_{t}\right)}, \notag\\
&\propto p_{\bm{\theta}}(\bm{x}_{t-1} | \bm{x}_t) \cdot p_{\bm{\theta}}(\bm{y}_{t-1} | \bm{x}_{t-1}).\label{x-from-y-and-x}
\end{align}
The process of sampling $\bm{x}_{t-1}$ based on $\bm{x}_t$ and $\bm{y}_{t-1}$ follows the probability distribution described in \eqref{x-from-y-and-x}. As outlined in \cite{box2011bayesian}, Monte Carlo sampling is employed to estimate the desired quantity, as is shown in Fig.~\ref{fig_sampling}. In Section \ref{sec-2-b}, it is described that a trained DM $\mu_{\bm{\theta}}$ can sample $\bm{x}_{t-1}$ from $\bm{x}_t$ using the following distribution.
\begin{align}
\mathcal{N}(\bm{x}_{t-1} | \mu_{\bm{\theta}}(\bm{x}_t, t), \sigma_t^2).\label{sample-x-t-1}
\end{align}
The procedure begins by sampling $M$ instances of $\bm{x}_{t-1}^{i}$, where $i \in \{1, 2, \cdots, M\}$, according to the distribution given by \eqref{sample-x-t-1}. Next, the log-probability $p_{\bm{\theta}}(\bm{x}_{t-1}^{i} | \bm{x}_t)$ is approximated using the following expression.
\begin{align}
\log p_{\bm{\theta}}(\bm{x}_{t-1}^{i} | \bm{x}_t) \propto -\frac{\|\bm{x}_{t-1}^{i} - \mu_{\bm{\theta}}(\bm{x}_t, t)\|^2}{2 \sigma_t^2}.
\end{align}
Since $\bm{x}_t$ is given, there is no need to include the superscript on $\bm{x}_t$. Given $\bm{y}_{t-1}$, which is generated as described in Section \ref{section4-2-1}, and the sequence $\{\bm{x}_{t-1}^{i}\}_{i=1}^{M}$, the log-probability $p_{\bm{\theta}}(\bm{y}_{t-1} | \bm{x}_{t-1}^{i})$ is expressed as follows.
\begin{align}
\log p_{\bm{\theta}}(\bm{y}_{t-1} | \bm{x}_{t-1}^{i}) \propto -\frac{\|\bm{y}_{t-1} - \bm{A} \bm{x}_{t-1}^{i}\|^2}{2 c_t^2 \sigma_t^2}.
\end{align}
Thus, the combined log-probability is as follows.
\begin{align}
\log p_{\bm{\theta}}(\bm{x}_{t-1}^{i} | \bm{x}_t, \bm{y}_{t-1}) &\propto \log p_{\bm{\theta}}(\bm{x}_{t-1}^{i} | \bm{x}_t) \notag\\
&\qquad\qquad + \log p_{\bm{\theta}}(\bm{y}_{t-1} | \bm{x}_{t-1}^{i}),\\
&\propto -\left(\frac{\|\bm{x}_{t-1}^{i} - \mu_{\bm{\theta}}(\bm{x}_t, t)\|^2}{2 \sigma_t^2}\right. \notag \\
&\left. \qquad\qquad+ \frac{\|\bm{y}_{t-1}- \bm{A} \bm{x}_{t-1}^{i}\|^2}{2 c_t^2 \sigma_t^2}\right).
\end{align}
To finalize the sampling process, each of $\bm{x}_{t-1}^{i}$ is sampled by probability $p_{\bm{\theta}}^{sample}(\bm{x}_{t-1}^{i}|\bm{x}_{t},\bm{y}_{t-1})$ as follows.
\begin{align}
    &p_{\bm{\theta}}^{sample}(\bm{x}_{t-1}^{i}|\bm{x}_{t},\bm{y}_{t-1})=\frac{p_{\bm{\theta}}^{wgt}(\bm{x}_{t-1}^{i}|\bm{x}_{t},\bm{y}_{t-1})}{\sum_{j=1}^{M}p_{\bm{\theta}}^{wgt}(\bm{x}_{t-1}^{j}|\bm{x}_{t},\bm{y}_{t-1})},\\
    &p_{\bm{\theta}}^{wgt}(\bm{x}_{t-1}^{i}|\bm{x}_{t},\bm{y}_{t-1})=e^{-\left(\frac{\|\bm{x}_{t-1}^{i} - \mu_{\bm{\theta}}(\bm{x}_t, t)\|^2}{2 \sigma_t^2}+\frac{\|\bm{y}_{t-1}- \bm{A} \bm{x}_{t-1}^{i}\|^2}{2 c_t^2 \sigma_t^2}\right)}.
\end{align}
The total sampling procedure from the $\bm{y}_{0}$ to obtain the $\bm{x}_{0}$ is shown in Fig.~\ref{fig-inference-sequence}.

\begin{remark}
    The RadioDiff-Inverse is a training-free framework, where the DM pre-trained on Imagenet without any task-specific fine-tune or post-training is used as the backbone NN of $\mu_{\bm{\theta}}$. The detail of how to train a DM can be found in \cite{LDM}.
\end{remark}

\section{Experiments}
\subsection{Dataset}
This study employs the DPM (Dynamic Pathloss Map) dataset from the RM pathloss construction challenge to evaluate our proposed method. The dataset encompasses 700 unique radio maps, each containing distinct geographic information and building configurations. Each map incorporates 80 transmitter locations with corresponding ground truth measurements and features between 50 and 150 buildings. We designate 200 maps for testing purposes, ensuring no overlap in terrain information between training and test sets to maintain evaluation integrity.

The dataset represents diverse urban environments, sourced from OpenStreetMap, including major metropolitan areas such as Ankara, Berlin, Glasgow, Ljubljana, London, and Tel Aviv. Environmental parameters are standardized across all maps: both transmitter and receiver heights are fixed at 1.5 meters, while building heights are uniformly set at 25 meters. Each map is discretized into a 256 × 256 pixel binary morphological image, where each pixel represents a one-square-meter area. The binary values denote building presence (1) or absence (0), providing a precise representation of the urban landscape.

Transmitter positions are encoded using two-dimensional numerical coordinates and represented in morphological images through binary indicators, where the transmitter's location is marked as 1 and all other positions as 0. The transmission parameters are standardized with 23 dBm transmitter power and 5.9 GHz carrier frequency. Ground truth radio maps (RMs) are generated using Maxwell's equations, accounting for electromagnetic ray reflection and diffraction phenomena. The dataset provides two variants of ground truth maps: Static Radio Maps (SRM), which consider only the electromagnetic interactions with static building structures, and Dynamic Radio Maps (DRM), which incorporate both static building effects and the influence of randomly distributed vehicles along roadways.

This comprehensive dataset enables rigorous evaluation of radio map reconstruction methods under diverse urban scenarios and propagation conditions, providing a robust foundation for assessing the performance of our proposed approach.


\begin{figure*}[t]
\captionsetup{font={small}, skip=16pt}
\scriptsize
\begin{tabular}{ccc}
\hspace{-0.5cm}
\begin{adjustbox}{valign=t}
\begin{tabular}{c}
\end{tabular}
\end{adjustbox}
\begin{adjustbox}{valign=t}
\begin{tabular}{cccccccc}
\includegraphics[width=0.16\linewidth]{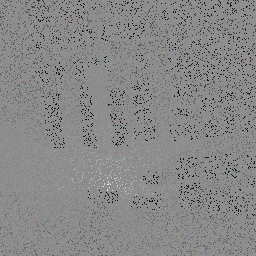} \hspace{-4mm} &
\includegraphics[width=0.16\linewidth]{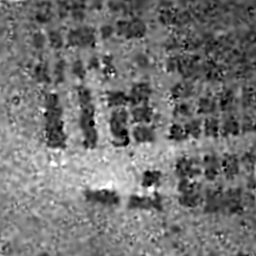}   \hspace{-4mm} &
\includegraphics[width=0.16\linewidth]{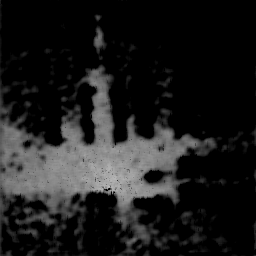}  \hspace{-4mm} &
\includegraphics[width=0.16\linewidth]{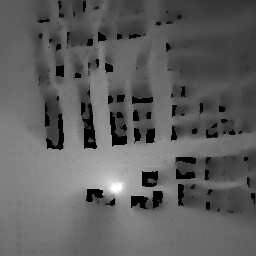}  \hspace{-4mm} &
\includegraphics[width=0.16\linewidth]{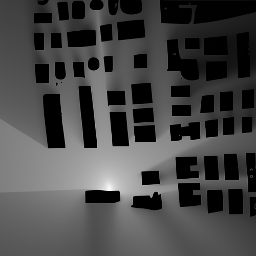}  \hspace{-4mm} &
\includegraphics[width=0.16\linewidth]{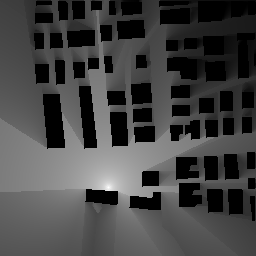}  \hspace{-4mm} &
\end{tabular}
\end{adjustbox}
\vspace{0.1mm}
\\
\hspace{-0.5cm}
\begin{adjustbox}{valign=t}
\begin{tabular}{c}
\end{tabular}
\end{adjustbox}
\begin{adjustbox}{valign=t}
\begin{tabular}{cccccccc}
\includegraphics[width=0.16\linewidth]{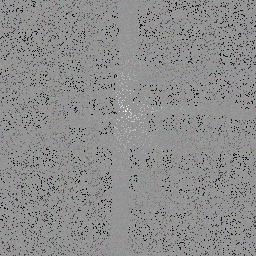} \hspace{-4mm} &
\includegraphics[width=0.16\linewidth]{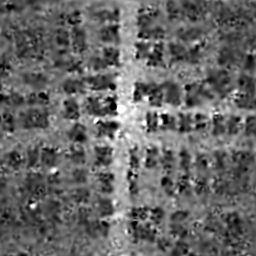}   \hspace{-4mm} &
\includegraphics[width=0.16\linewidth]{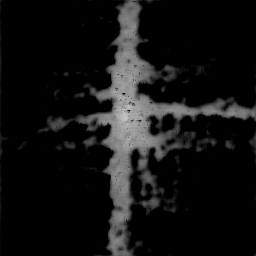}  \hspace{-4mm} &
\includegraphics[width=0.16\linewidth]{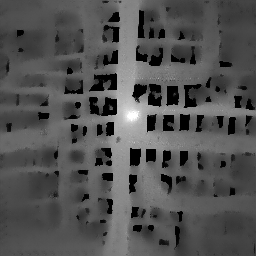}  \hspace{-4mm} &
\includegraphics[width=0.16\linewidth]{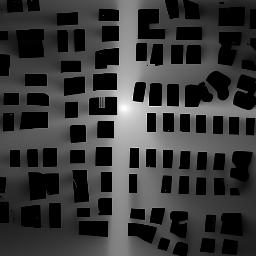}  \hspace{-4mm} &
\includegraphics[width=0.16\linewidth]{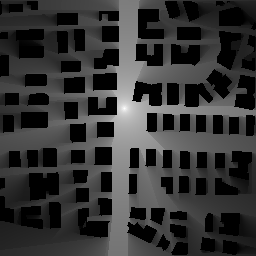}  \hspace{-4mm} &
\end{tabular}
\end{adjustbox}
\vspace{0.1mm}
\\
\hspace{-0.5cm}
\begin{adjustbox}{valign=t}
\begin{tabular}{c}
\end{tabular}
\end{adjustbox}
\begin{adjustbox}{valign=t}
\begin{tabular}{cccccccc}
\includegraphics[width=0.16\linewidth]{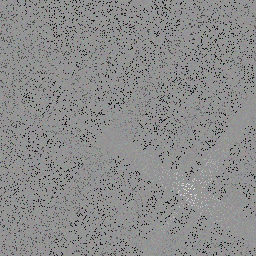} \hspace{-4mm} &
\includegraphics[width=0.16\linewidth]{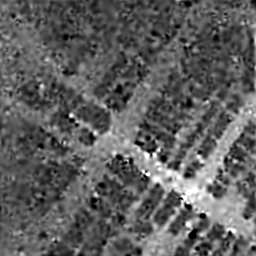}   \hspace{-4mm} &
\includegraphics[width=0.16\linewidth]{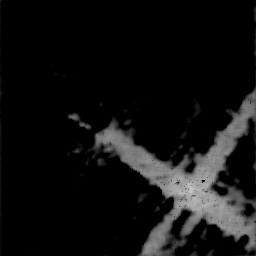}  \hspace{-4mm} &
\includegraphics[width=0.16\linewidth]{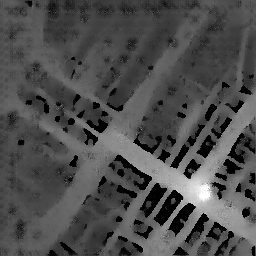}  \hspace{-4mm} &
\includegraphics[width=0.16\linewidth]{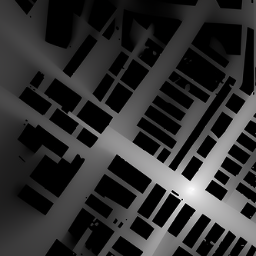}  \hspace{-4mm} &
\includegraphics[width=0.16\linewidth]{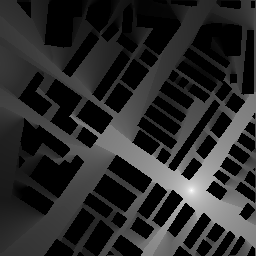}  \hspace{-4mm} &
\end{tabular}
\end{adjustbox}
\vspace{0.1mm}
\\
\hspace{-0.55cm}
\begin{adjustbox}{valign=t}
\begin{tabular}{c}
\end{tabular}
\end{adjustbox}
\begin{adjustbox}{valign=t}
\begin{tabular}{cccccccc}
\includegraphics[width=0.16\linewidth]{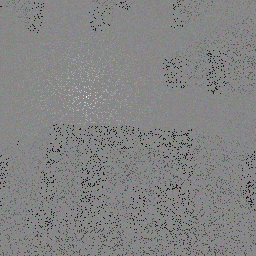} \hspace{-4mm} &
\includegraphics[width=0.16\linewidth]{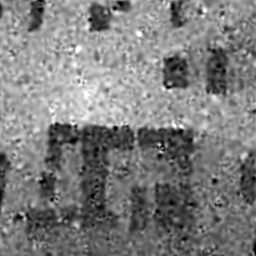}   \hspace{-4mm} &
\includegraphics[width=0.16\linewidth]{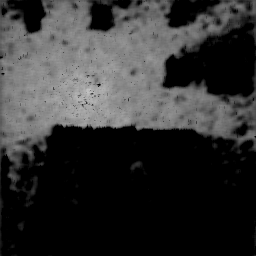}  \hspace{-4mm} &
\includegraphics[width=0.16\linewidth]{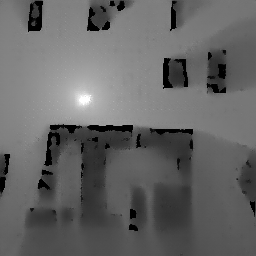}  \hspace{-4mm} &
\includegraphics[width=0.16\linewidth]{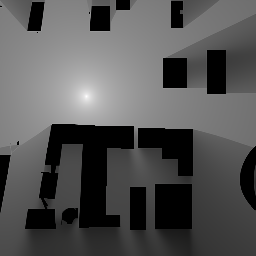}  \hspace{-4mm} &
\includegraphics[width=0.16\linewidth]{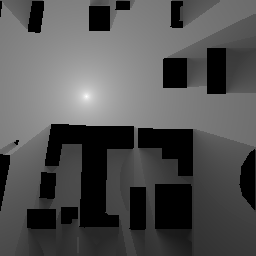}  \hspace{-4mm} &
\\
Input \hspace{-4mm} &
Interpolation \hspace{-4mm} &
RadioUNet \hspace{-4mm} &
RME-GAN \hspace{-4mm} &
Radiodiff-Inverse (Ours) \hspace{-4mm} &
Ground Truth
\\
\end{tabular}
\end{adjustbox}
\end{tabular}
\caption{The comparision of constructed RM between different methods}
\label{fig:r2}
\vspace{-12pt}
\end{figure*}

\subsection{Evaluation Framework}
We establish a comprehensive evaluation framework that assesses the performance of our proposed diffusion model across three critical dimensions: reconstruction quality, source localization accuracy, and electromagnetic interaction modeling. This framework enables systematic assessment of our method's effectiveness in real-world electromagnetic sensing applications.
\subsubsection{Reconstruction Quality Metrics}
To quantify the fidelity of reconstructed radio maps, we employ four complementary metrics that provide comprehensive quality assessment. The Peak Signal-to-Noise Ratio (PSNR) measures the logarithmic ratio between maximum possible signal power and noise power, defined as:
\begin{equation}
    \text{PSNR} = 10 \cdot \log_{10}\left(\frac{\text{MAX}^2}{\text{MSE}}\right)
\end{equation}
where MSE denotes mean squared error and MAX represents the maximum signal value.

The Structural Similarity Index (SSIM) evaluates perceptual similarity through luminance, contrast, and structure comparison:
\begin{equation}
    \text{SSIM}(x,y) = \frac{(2\mu_x\mu_y + C_1)(2\sigma_{xy} + C_2)}{(\mu_x^2 + \mu_y^2 + C_1)(\sigma_x^2 + \sigma_y^2 + C_2)}
\end{equation}
where $\mu$, $\sigma$ represent mean and variance respectively, and $C_1$, $C_2$ are stabilization constants.

For scale-invariant error assessment, we utilize the Normalized Mean Square Error (NMSE):
\begin{equation}
    \text{NMSE} = \frac{\sum_{i=1}^{N}(x_i - \hat{x}_i)^2}{\sum_{i=1}^{N}x_i^2}
\end{equation}

Additionally, the Root Mean Square Error (RMSE) quantifies absolute reconstruction error in physical units:
\begin{equation}
    \text{RMSE} = \sqrt{\frac{1}{N}\sum_{i=1}^{N}(x_i - \hat{x}_i)^2}
\end{equation}
\subsubsection{Source Localization Metrics}
For evaluating emission source localization accuracy, we introduce two specialized metrics. The Source Position Error (SPE) measures the Euclidean distance (in pixels) between predicted and actual emission sources. Our experiments demonstrate an average SPE of 0.85 pixels under challenging conditions (mask rate: 0.70, noise level: 0.070). The Source Region Quality (SRQ) evaluates reconstruction accuracy in the vicinity of emission sources through region-specific calculations of PSNR, SSIM, NMSE, and RMSE. Under standard test conditions, we achieve exceptional accuracy in source-adjacent regions, with SRQ-PSNR of 47.25 dB and SRQ-SSIM of 0.999.

\subsubsection{Electromagnetic Interaction Metrics}
To assess the model's capability in capturing building-induced electromagnetic effects, we define the Building Impact Assessment (BIA) metric. This specialized measure quantifies reconstruction accuracy specifically in building-affected regions, achieving BIA-PSNR of 40.99 dB and BIA-SSIM of 0.958, demonstrating effective modeling of electromagnetic interactions with structural elements.

Our evaluation framework systematically assesses performance across diverse operational scenarios, including varying building presence conditions, different sampling strategies (structured vs. random), mask ratios (0.50 to 0.95), and noise levels (0.01 to 0.09). This comprehensive approach ensures thorough performance characterization across scenarios representative of real-world electromagnetic sensing applications.

\subsection{Implementation Details}
We implement our radio map reconstruction framework using PyTorch, leveraging a pre-trained diffusion model initially developed for ImageNet classification. The implementation balances computational efficiency with reconstruction fidelity, making it suitable for both offline analysis and semi-real-time applications in electromagnetic environment assessment.

\subsubsection{Model Architecture}
The core architecture employs a modified U-Net backbone designed for optimal radio map reconstruction. The network is configured with a base channel dimension of 256 and incorporates two residual blocks per resolution level. Operating at an input resolution of 256$\times$256 pixels, the architecture integrates multi-head attention mechanisms at 32$\times$32, 16$\times$16, and 8$\times$8 resolutions, utilizing 4 attention heads with 64 channels per head. Key architectural enhancements include scale-shift normalization with residual connections, adaptive feature fusion across resolution levels, and progressive upsampling with skip connections.

\subsubsection{Diffusion Process Configuration}
The diffusion process employs a linear noise schedule with DDIM sampling method\cite{song2020denoising}, configured for 1000 timesteps. The process utilizes $\epsilon$-prediction with learned range variance for optimal noise prediction. Training optimizations incorporate gradient clipping during denoising, preservation of original timestep scaling, and adaptive learning rate scheduling. This configuration ensures stable training and efficient sampling while maintaining high reconstruction quality.

\subsubsection{Computational Environment}
Our experiments are conducted on NVIDIA H100 GPUs with 80GB HBM3 memory per unit, supported by 512GB DDR5 system memory. The implementation achieves an average reconstruction time of 100 seconds per 256$\times$256 map, with the capability to process 16 maps simultaneously. Peak memory usage is maintained at 45GB, demonstrating efficient resource utilization. The modular architecture design facilitates easy adaptation to different input resolutions and environmental configurations while maintaining consistent performance characteristics.

\subsection{Comparative Analysis}
We evaluate our RadioDiff-inverse model against state-of-the-art approaches in radio map reconstruction, including both deep learning-based and traditional methods. For fairness, all compared methods use identical training and testing datasets.

\subsubsection{Comparison Methods}
\begin{itemize}
    \item \textbf{RadioUNet}~\cite{levie2021radiounet}: A convolutional U-Net architecture that learns environmental characteristics for radio map reconstruction. As a supervised learning approach, RadioUNet represents the current standard in non-generative neural network solutions.
    
    \item \textbf{RME-GAN}~\cite{zhang2023rme}: A conditional GAN architecture that generates radio maps from environmental features. While the original implementation incorporates pathloss measurements, our comparison uses only environmental features to maintain consistency with our method's inputs.
    
    \item \textbf{Kriging Interpolation}: A geostatistical technique that estimates unknown values based on spatial correlation of sampled points. We implement ordinary Kriging with exponential variogram models.
    
\end{itemize}

\subsubsection{Evaluation Protocol}
All neural network models were trained until convergence with early stopping based on validation loss. We maintain consistent hyperparameter optimization procedures across all learning-based methods. The evaluation metrics include:

\begin{itemize}
    \item Peak Signal-to-Noise Ratio (PSNR)
    \item Structural Similarity Index (SSIM)
    \item Normalized Mean Square Error (NMSE)
\end{itemize}

For statistical significance, we report mean values across 5 independent runs with different random seeds, along with 95\% confidence intervals.
\subsection{Experimental Results and Analysis}

We conduct comprehensive experiments to evaluate our proposed inverse method for radio map reconstruction. The evaluation encompasses four key aspects: 1) the impact of building information on reconstruction accuracy, 2) the effectiveness of different sampling strategies, 3) the influence of mask ratios, and 4) the robustness against noise. Tables \ref{tab:random_mask_noise01}-\ref{tab:random_mask_noise09} present the quantitative results across these different experimental configurations.

We conduct comprehensive experiments to evaluate our proposed inverse method for radio map reconstruction. The evaluation encompasses four key aspects: 1) the impact of building information on reconstruction accuracy, 2) the effectiveness of different sampling strategies, 3) the influence of mask ratios, and 4) the robustness against noise. Tables \ref{tab:random_mask_noise01}-\ref{tab:random_mask_noise09} present the quantitative results across these different experimental configurations.

\begin{table}[!t]
    \centering
    \small
    \caption{Performance Comparison with Random Pixel-wise Masking (Noise Level = 0.01)}
    \label{tab:random_mask_noise01}
    \setlength{\tabcolsep}{3pt}
    \begin{tabular}{|l|c|c|c|c|}
    \hline
    \multirow{2}{*}{Sampling Setting} & \multicolumn{4}{c|}{Method} \\
    \cline{2-5}
     & Ours & RadioUNet & RME-GAN & Interp. \\
    \hline
    \multicolumn{5}{|c|}{PSNR (↑)} \\
    \hline
    Aware (70\% Missing) & {\color[HTML]{9A0000} \textbf{34.55}} & 22.17 & {\color[HTML]{00009B} \underline{23.97}} & 22.40 \\
    Aware (80\% Missing) & {\color[HTML]{9A0000} \textbf{32.57}} & 22.05 & {\color[HTML]{00009B} \underline{26.22}} & 22.18 \\
    Aware (90\% Missing) & {\color[HTML]{9A0000} \textbf{27.83}} & 22.10 & {\color[HTML]{00009B} \underline{26.03}} & 22.40 \\
    Unaware (70\% Missing) & {\color[HTML]{9A0000} \textbf{33.91}} & 22.17 & {\color[HTML]{00009B} \underline{23.97}} & 22.40 \\
    Unaware (80\% Missing) & {\color[HTML]{9A0000} \textbf{31.93}} & 22.05 & {\color[HTML]{00009B} \underline{26.22}} & 22.18 \\
    Unaware (90\% Missing) & {\color[HTML]{9A0000} \textbf{27.19}} & 22.10 & {\color[HTML]{00009B} \underline{26.03}} & 22.40 \\
    \hline
    \multicolumn{5}{|c|}{SSIM (↑)} \\
    \hline
    Aware (70\% Missing) & {\color[HTML]{9A0000} \textbf{0.9847}} & {\color[HTML]{00009B} \underline{0.9185}} & 0.7816 & 0.7914 \\
    Aware (80\% Missing) & {\color[HTML]{9A0000} \textbf{0.9765}} & {\color[HTML]{00009B} \underline{0.9172}} & 0.7953 & 0.7855 \\
    Aware (90\% Missing) & {\color[HTML]{9A0000} \textbf{0.9420}} & {\color[HTML]{00009B} \underline{0.9173}} & 0.7622 & 0.7942 \\
    Unaware (70\% Missing) & {\color[HTML]{9A0000} \textbf{0.9836}} & {\color[HTML]{00009B} \underline{0.9185}} & 0.7816 & 0.7914 \\
    Unaware (80\% Missing) & {\color[HTML]{9A0000} \textbf{0.9754}} & {\color[HTML]{00009B} \underline{0.9172}} & 0.7953 & 0.7855 \\
    Unaware (90\% Missing) & {\color[HTML]{9A0000} \textbf{0.9409}} & {\color[HTML]{00009B} \underline{0.9173}} & 0.7622 & 0.7942 \\
    \hline
    \multicolumn{5}{|c|}{NMSE (↓)} \\
    \hline
    Aware (70\% Missing) & {\color[HTML]{9A0000} \textbf{0.0050}} & {\color[HTML]{00009B} \underline{0.0116}} & 0.0422 & 0.0699 \\
    Aware (80\% Missing) & {\color[HTML]{9A0000} \textbf{0.0078}} & {\color[HTML]{00009B} \underline{0.0118}} & 0.0242 & 0.0691 \\
    Aware (90\% Missing) & {\color[HTML]{00009B} \underline{0.0208}} & {\color[HTML]{9A0000} \textbf{0.0118}} & 0.0244 & 0.0700 \\
    Unaware (70\% Missing) & {\color[HTML]{9A0000} \textbf{0.0055}} & {\color[HTML]{00009B} \underline{0.0116}} & 0.0422 & 0.0699 \\
    Unaware (80\% Missing) & {\color[HTML]{9A0000} \textbf{0.0083}} & {\color[HTML]{00009B} \underline{0.0118}} & 0.0242 & 0.0691 \\
    Unaware (90\% Missing) & {\color[HTML]{00009B} \underline{0.0213}} & {\color[HTML]{9A0000} \textbf{0.0118}} & 0.0244 & 0.0700 \\
    \hline
    \end{tabular}
    \vspace{-2mm}
\end{table}

\begin{table}[!t]
    \centering
    \small
    \caption{Performance Comparison with Random Pixel-wise Masking (Noise Level = 0.05)}
    \label{tab:random_mask_noise05}
    \setlength{\tabcolsep}{3pt}
    \begin{tabular}{|l|c|c|c|c|}
    \hline
    \multirow{2}{*}{Sampling Setting} & \multicolumn{4}{c|}{Method} \\
    \cline{2-5}
     & Ours & RadioUNet & RME-GAN & Interp. \\
    \hline
    \multicolumn{5}{|c|}{PSNR (↑)} \\
    \hline
    Aware (70\% Missing) & {\color[HTML]{9A0000} \textbf{35.29}} & 22.19 & {\color[HTML]{00009B} \underline{23.67}} & 22.12 \\
    Aware (80\% Missing) & {\color[HTML]{9A0000} \textbf{32.16}} & 22.22 & {\color[HTML]{00009B} \underline{24.89}} & 22.26 \\
    Aware (90\% Missing) & {\color[HTML]{9A0000} \textbf{27.44}} & 22.04 & {\color[HTML]{00009B} \underline{25.61}} & 22.11 \\
    Unaware (70\% Missing) & {\color[HTML]{9A0000} \textbf{34.65}} & 22.19 & {\color[HTML]{00009B} \underline{23.67}} & 22.12 \\
    Unaware (80\% Missing) & {\color[HTML]{9A0000} \textbf{31.52}} & 22.22 & {\color[HTML]{00009B} \underline{24.89}} & 22.26 \\
    Unaware (90\% Missing) & {\color[HTML]{9A0000} \textbf{26.80}} & 22.04 & {\color[HTML]{00009B} \underline{25.61}} & 22.11 \\
    \hline
    \multicolumn{5}{|c|}{SSIM (↑)} \\
    \hline
    Aware (70\% Missing) & {\color[HTML]{9A0000} \textbf{0.9840}} & {\color[HTML]{00009B} \underline{0.9187}} & 0.7791 & 0.7723 \\
    Aware (80\% Missing) & {\color[HTML]{9A0000} \textbf{0.9748}} & {\color[HTML]{00009B} \underline{0.9183}} & 0.7868 & 0.7743 \\
    Aware (90\% Missing) & {\color[HTML]{9A0000} \textbf{0.9376}} & {\color[HTML]{00009B} \underline{0.9167}} & 0.7559 & 0.7689 \\
    Unaware (70\% Missing) & {\color[HTML]{9A0000} \textbf{0.9829}} & {\color[HTML]{00009B} \underline{0.9187}} & 0.7791 & 0.7723 \\
    Unaware (80\% Missing) & {\color[HTML]{9A0000} \textbf{0.9737}} & {\color[HTML]{00009B} \underline{0.9183}} & 0.7868 & 0.7743 \\
    Unaware (90\% Missing) & {\color[HTML]{9A0000} \textbf{0.9365}} & {\color[HTML]{00009B} \underline{0.9167}} & 0.7559 & 0.7689 \\
    \hline
    \multicolumn{5}{|c|}{NMSE (↓)} \\
    \hline
    Aware (70\% Missing) & {\color[HTML]{9A0000} \textbf{0.0043}} & {\color[HTML]{00009B} \underline{0.0116}} & 0.0424 & 0.0683 \\
    Aware (80\% Missing) & {\color[HTML]{9A0000} \textbf{0.0083}} & {\color[HTML]{00009B} \underline{0.0116}} & 0.0320 & 0.0679 \\
    Aware (90\% Missing) & {\color[HTML]{00009B} \underline{0.0223}} & {\color[HTML]{9A0000} \textbf{0.0119}} & 0.0273 & 0.0714 \\
    Unaware (70\% Missing) & {\color[HTML]{9A0000} \textbf{0.0048}} & {\color[HTML]{00009B} \underline{0.0116}} & 0.0424 & 0.0683 \\
    Unaware (80\% Missing) & {\color[HTML]{9A0000} \textbf{0.0088}} & {\color[HTML]{00009B} \underline{0.0116}} & 0.0320 & 0.0679 \\
    Unaware (90\% Missing) & {\color[HTML]{00009B} \underline{0.0228}} & {\color[HTML]{9A0000} \textbf{0.0119}} & 0.0273 & 0.0714 \\
    \hline
    \end{tabular}
    \vspace{-2mm}
\end{table}

\begin{table}[!t]
    \centering
    \small
    \caption{Performance Comparison with Random Pixel-wise Masking (Noise Level = 0.09)}
    \label{tab:random_mask_noise09}
    \setlength{\tabcolsep}{3pt}
    \begin{tabular}{|l|c|c|c|c|}
    \hline
    \multirow{2}{*}{Sampling Setting} & \multicolumn{4}{c|}{Method} \\
    \cline{2-5}
     & Ours & RadioUNet & RME-GAN & Interp. \\
    \hline
    \multicolumn{5}{|c|}{PSNR (↑)} \\
    \hline
    Aware (70\% Missing) & {\color[HTML]{9A0000} \textbf{33.35}} & {\color[HTML]{00009B} \underline{21.99}} & 21.82 & 21.70 \\
    Aware (80\% Missing) & {\color[HTML]{9A0000} \textbf{31.82}} & 21.84 & {\color[HTML]{00009B} \underline{23.52}} & 22.03 \\
    Aware (90\% Missing) & {\color[HTML]{9A0000} \textbf{28.17}} & 21.82 & {\color[HTML]{00009B} \underline{25.64}} & 21.99 \\
    Unaware (70\% Missing) & {\color[HTML]{9A0000} \textbf{32.71}} & {\color[HTML]{00009B} \underline{21.99}} & 21.82 & 21.70 \\
    Unaware (80\% Missing) & {\color[HTML]{9A0000} \textbf{31.18}} & 21.84 & {\color[HTML]{00009B} \underline{23.52}} & 22.03 \\
    Unaware (90\% Missing) & {\color[HTML]{9A0000} \textbf{27.53}} & 21.82 & {\color[HTML]{00009B} \underline{25.64}} & 21.99 \\
    \hline
    \multicolumn{5}{|c|}{SSIM (↑)} \\
    \hline
    Aware (70\% Missing) & {\color[HTML]{00009B} \underline{0.8483}} & {\color[HTML]{9A0000} \textbf{0.9159}} & 0.7610 & 0.7436 \\
    Aware (80\% Missing) & {\color[HTML]{00009B} \underline{0.8771}} & {\color[HTML]{9A0000} \textbf{0.9145}} & 0.7730 & 0.7501 \\
    Aware (90\% Missing) & {\color[HTML]{00009B} \underline{0.8501}} & {\color[HTML]{9A0000} \textbf{0.9144}} & 0.7417 & 0.7449 \\
    Unaware (70\% Missing) & {\color[HTML]{00009B} \underline{0.8472}} & {\color[HTML]{9A0000} \textbf{0.9159}} & 0.7610 & 0.7436 \\
    Unaware (80\% Missing) & {\color[HTML]{00009B} \underline{0.8760}} & {\color[HTML]{9A0000} \textbf{0.9145}} & 0.7730 & 0.7501 \\
    Unaware (90\% Missing) & {\color[HTML]{00009B} \underline{0.8490}} & {\color[HTML]{9A0000} \textbf{0.9144}} & 0.7417 & 0.7449 \\
    \hline
    \multicolumn{5}{|c|}{NMSE (↓)} \\
    \hline
    Aware (70\% Missing) & {\color[HTML]{9A0000} \textbf{0.0057}} & {\color[HTML]{00009B} \underline{0.0121}} & 0.0657 & 0.0771 \\
    Aware (80\% Missing) & {\color[HTML]{9A0000} \textbf{0.0089}} & {\color[HTML]{00009B} \underline{0.0124}} & 0.0467 & 0.0740 \\
    Aware (90\% Missing) & {\color[HTML]{00009B} \underline{0.0190}} & {\color[HTML]{9A0000} \textbf{0.0123}} & 0.0262 & 0.0714 \\
    Unaware (70\% Missing) & {\color[HTML]{9A0000} \textbf{0.0062}} & {\color[HTML]{00009B} \underline{0.0121}} & 0.0657 & 0.0771 \\
    Unaware (80\% Missing) & {\color[HTML]{9A0000} \textbf{0.0094}} & {\color[HTML]{00009B} \underline{0.0124}} & 0.0467 & 0.0740 \\
    Unaware (90\% Missing) & {\color[HTML]{00009B} \underline{0.0195}} & {\color[HTML]{9A0000} \textbf{0.0123}} & 0.0262 & 0.0714 \\
    \hline
    \end{tabular}
    \vspace{-2mm}
\end{table}

\begin{figure}
    \centering
    \includegraphics[width=1.0\linewidth]{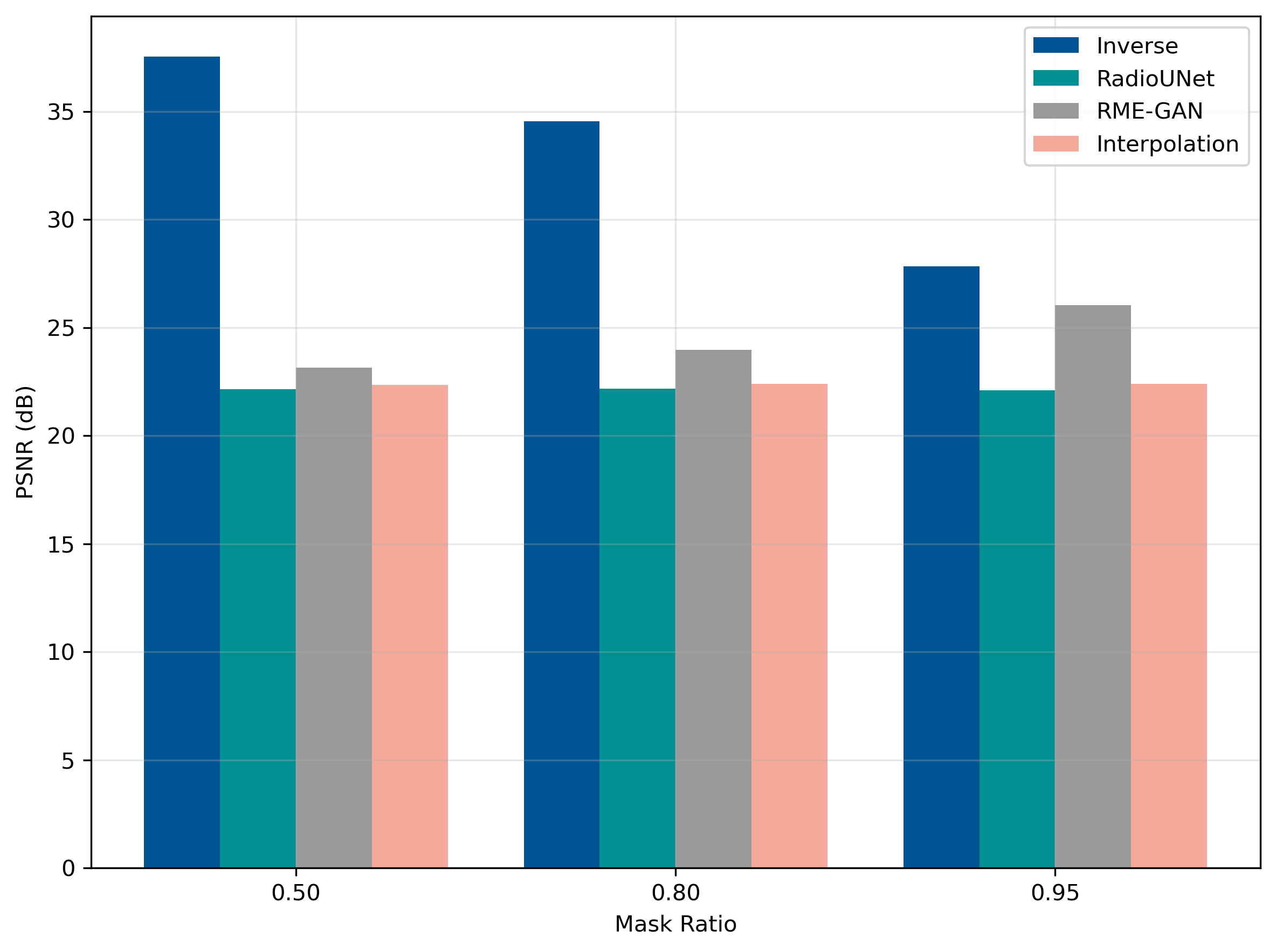}
    \caption{Performance Comparision of All Methods with random masking and building information}
    \label{fig:cond_random_psnr}
\end{figure}

\begin{figure}
    \centering
    \includegraphics[width=1.0\linewidth]{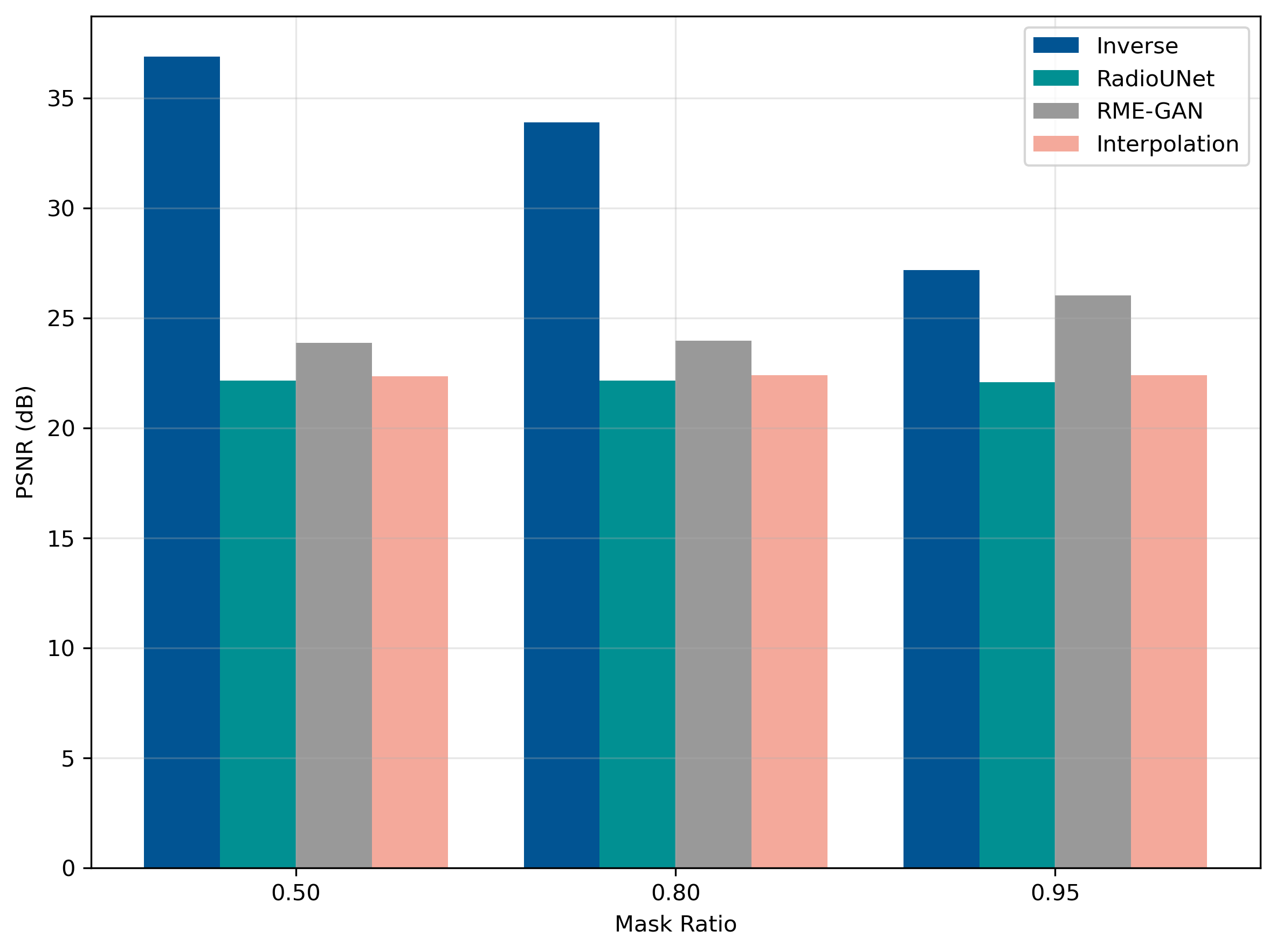}
    \caption{Performance Comparision of All Methods with random masking without building information}
    \label{fig:uncond_random_psnr}
\end{figure}

\subsubsection{Impact of Building Information}
The experimental results demonstrate that building information significantly enhances radio map reconstruction accuracy. Our analysis reveals a consistent performance advantage of scenario-aware generation (with building information) over scenario-unaware approaches across all test configurations. At low mask ratios (70\%), scenario-aware generation achieves superior performance with a PSNR of 34.55dB and SSIM of 0.9847, compared to scenario-unaware generation's 33.91dB PSNR and 0.9836 SSIM. This performance gap becomes more pronounced as the reconstruction task grows more challenging with higher mask ratios. At 80\% masking, the scenario-aware approach maintains a PSNR of 32.57dB, outperforming its unaware counterpart by 0.64dB (31.93dB). The advantage extends to structural preservation metrics, with scenario-aware generation consistently achieving higher SSIM values (0.9765 vs. 0.9754 at 80\% masking), indicating better preservation of critical radio map features.

\subsubsection{Effect of Sampling Strategy}
The comparison between random pixel-wise masking and structured sampling reveals distinct characteristics that influence reconstruction quality. Our method demonstrates remarkable resilience under random pixel-wise masking, maintaining high PSNR values even at challenging mask ratios. The performance advantage is particularly evident in scenario-aware settings, where our approach achieves 35.29dB PSNR at 70\% masking with moderate noise (0.05), significantly outperforming both RadioUNet (22.19dB) and RME-GAN (23.67dB). The SSIM metrics further corroborate this superiority, with our method achieving 0.9840 compared to RadioUNet's 0.9187 and RME-GAN's 0.7791 under the same conditions.

\subsubsection{Mask Ratio Analysis}
The impact of mask ratio reveals important insights into the robustness and scalability of different reconstruction approaches. As mask ratios increase from 70\% to 90\%, all methods exhibit performance degradation, but with varying degrees of resilience. Our inverse method demonstrates remarkable stability, maintaining PSNR above 27dB even at 90\% masking in scenario-aware settings. The performance degradation follows a consistent pattern, with PSNR decreasing from 34.55dB at 70\% to 27.83dB at 90\% masking under noise level 0.01. Notably, while RME-GAN shows competitive performance at high mask ratios (achieving 26.03dB PSNR at 90\% masking), our method maintains superior structural preservation as evidenced by consistently higher SSIM values (0.9420 vs. 0.7622).

\subsubsection{Noise Level Impact}
The analysis of noise impact reveals crucial insights into the robustness of different reconstruction methods. Our inverse method demonstrates exceptional resilience across varying noise levels (0.01-0.09), maintaining strong performance metrics even under challenging conditions. At moderate noise (0.05), the method achieves 35.29dB PSNR and 0.9840 SSIM with 70\% masking in scenario-aware settings. Even under high noise conditions (0.09), the performance remains robust, with PSNR values of 33.35dB and 31.82dB at 70\% and 80\% masking, respectively. This represents a significant advantage over competing methods, with RadioUNet achieving only 21.99dB PSNR under similar conditions. The NMSE metrics further validate this robustness, with our method maintaining values below 0.01 in most scenarios, only slightly increasing to 0.0190 under the most challenging conditions (90\% masking, noise level 0.09).

\section{Conclusion}

This study has presented RadioDiff-Inverse, a diffusion-enhanced Bayesian framework for radio map reconstruction under environmental uncertainty and sparse noisy measurements. Two principal contributions have been established: (1) A Bayesian inverse problem formulation systematically addressing measurement and environmental uncertainties has been developed; (2) A training-free methodology leveraging pre-trained diffusion models for prior distribution learning has demonstrated effective cross-domain knowledge transfer. The framework's capability to infer environmental features from limited measurements has advanced integrated sensing and communication in semi-aware scenarios. Future research directions have been identified, including dynamic environment adaptation and integration with emerging wireless technologies. This work has established a new paradigm for addressing inverse problems in wireless communications through synergistic integration of generative models and Bayesian inference.

\bibliography{ref}
\bibliographystyle{IEEEtran}

\ifCLASSOPTIONcaptionsoff
  \newpage
\fi

\end{document}